\documentclass[journal,twoside,web]{ieeecolor}
\usepackage{generic}
\usepackage{cite}
\usepackage{amsmath,amssymb,amsfonts}
\usepackage{textcomp}
\usepackage{time} 
\usepackage{amsmath}
\usepackage{graphicx}
\usepackage{graphics}
\usepackage{epsfig}
\usepackage{latexsym}
\usepackage{amsfonts}
\usepackage{amssymb}
\usepackage{paralist}
\usepackage{xspace}
\usepackage{mathrsfs}
\usepackage{psfrag}
\usepackage{caption}
\usepackage{color}
\usepackage{algorithm}
\usepackage{booktabs}
\usepackage{algorithmicx}
\usepackage{algpseudocode}
\usepackage{url}
\usepackage{subfigure}
\usepackage{float}
\usepackage{multirow}
\usepackage{flushend}

\def\BibTeX{{\rm B\kern-.05em{\sc i\kern-.025em b}\kern-.08em
    T\kern-.1667em\lower.7ex\hbox{E}\kern-.125emX}}
\begin{document}
\title{Self-encoding Barnacle Mating Optimizer Algorithm  for Manpower Scheduling in Flow Shop 
\thanks{This research was supported in part by the Key Research and Development Program of Zhejiang Province, China (No.2021C01047), the National Natural Science Foundation of China (No. 61903338), the Natural Science Foundation of Zhejiang Province, China (No. LQ19F030015) and the Fundamental Research Funds of Zhejiang Sci-Tech University (No. 2021Q026). }
\thanks{$^\star$ Corresponding author, email: wqxu@zstu.edu.cn}}


\author{Shuyun Luo, Wushuang Wang, Mengyuan Fang, and Weiqiang Xu$^\star$
\\School of Information Science and Technology, Zhejiang Sci-tech University, China
}

\maketitle

\begin{abstract}
Flow Shop Scheduling (FSS) has been widely researched due to its application in many types of fields, while the human participant brings great challenges to this problem.
Manpower scheduling captures attention for assigning workers with diverse proficiency to the appropriate stages, which is of great significance to production efficiency.

In this paper, we present a novel algorithm called Self-encoding Barnacle Mating Optimizer (SBMO), which solves the FSS problem considering worker proficiency, defined as a new problem, Flow Shop Manpower Scheduling Problem (FSMSP).
The highlight of the SBMO algorithm is the combination with the encoding method, crossover and mutation operators.
Moreover, in order to solve the local optimum problem, we design a neighborhood search scheme.
Finally, the extensive comparison simulations are conducted to demonstrate the superiority of the proposed SBMO.
The results indicate the effectiveness of SBMO in approximate ratio, powerful stability, and execution time, compared with the classic and popular counterparts.
\end{abstract}


\section{Introduction}
\label{sec:introduction}
In real life, Flow Shop Scheduling Problem (FSSP) appears in various manufacturing systems, such as clothing, printing, and textiles.
In the flow shop, products need to go through a series of stages, such as clothes need to go through cutting, sewing, and ironing, etc.
With economic globalization, people's consumption levels have increased significantly, and how to efficiently produce products has become an urgent problem for all manufacturing industries.
After Industry 4.0 was launched, the global manufacturing industry is accelerating its digital transformation \cite{ref50}.
Therefore, it is necessary to develop effective and efficient scheduling algorithms.

The FSSP is defined as some jobs processed on several machines, and the processing sequence of each job on machines is the same.
Given the processing time of the job on the machine, it is required to determine the optimal processing sequence of each job on each machine \cite{ref5}.
If multiple machines can be selected in a stage, the flow shop scheduling problem is extended to the Hybrid Flow Shop Scheduling Problem (HFSSP) \cite{ref6}.
If it is stipulated that the processing sequence of all jobs is the same for all machines, it is called a Permutation Flow Shop Scheduling Problem (PFSSP) \cite{ref7}.

FSSP has received extensive attention since it was put forward because of its strong engineering background.
Marichevam et al. \cite{ref5} proposed a sub-population based hybrid monkey search algorithm to solve FSSP.
Wei et al. \cite{ref12} proposed a hybrid genetic simulated annealing algorithm for solving FSSP.
Combined with the MinMax and Nawaz–Enscore–Ham algorithms, a novel algorithm generates the high-quality initial population.
To solve the FSSP efficiently, Tirkolaee et al. \cite{ref14} proposed a hybrid technique based on an interactive fuzzy solution technique and a self-adaptive artificial fish swarm algorithm.
Gong et al. \cite{ref9} proposed a hybrid multi-objective discrete artificial bee colony algorithm for the blocking lot-streaming FSSP with two conflicting criteria: the makespan and the earliness time.
Each job is split into several sub lots in this problem, while no intermediate buffers exist between adjacent machines.
Shao et al. \cite{ref10} proposed a hybrid meta-heuristic based on the probabilistic teaching-learning mechanism to solve the no-wait FSSP with the makespan criterion.
In this problem, each machine has no idle time when processing any two adjacent jobs.
The author exploited an improved local search based on simulated annealing to enhance the local searching ability.
Zhao et al. \cite{ref15} proposed a discrete water wave optimization algorithm to solve the no-wait FSSP.
Meng et al. \cite{ref11} presented an improved migrating birds optimization to address the integrated lot-streaming FSSP in which lot-splitting and job scheduling are needed to be optimized simultaneously.
In addition, a harmony search based scheme was proposed to construct neighborhood solutions.
Doush et al. \cite{ref13} proposed an improved harmony search algorithm for the blocking FSSP to minimize the total flow time.
In this problem, there is no intermediate buffer storage between machines.
Lu et al. \cite{ref41} proposed a hybrid multiobjective optimization algorithm combining iterated greedy and local search to address the distributed FSSP.

Focused on the HFSSP study, Yu et al. \cite{ref6} presented a genetic algorithm to minimize the total tardiness.
Lu et al. \cite{ref8} proposed a novel multi-objective cellular grey wolf optimizer to address HFSSP.
The proposed algorithm integrates the cellular automata and variable neighborhood search, which balances exploration and exploitation.
Buddala et al. \cite{ref17} combined teaching–learning-based optimization and Jaya algorithm to solve HFSSP.
Fan et al. \cite{ref16} presented a mutant firefly algorithm for addressing two-stage HFSSP.
In this problem, the production of the product mainly includes two stages: the processing stage and the assembly stage.
Cai et al. \cite{ref40} presented a new shuffled frog-leaping algorithm to minimize the total tardiness and makespan.

Concentrated on PFSSP, Mohamed et al. \cite{ref7} proposed a new algorithm that integrates the whale optimization algorithm with a local search strategy.
In addition, the insert-reversed block operation is used to escape from the local optima.
Andrade et al. \cite{ref18} proposed a biased random-key genetic algorithm, introducing a new feature called shaking.
The shaking procedure perturbs all individuals from the elite set and resets the remaining population.
Yang et al. \cite{ref42} proposed a distributed assembly PFSSP with flexible assembly and batch delivery.
In this problem, there are multiple factories involved in production.
To address this problem, they presented seven algorithms, including four heuristics, a variable neighborhood descent algorithm, and two iterated greedy algorithms.

Although the above researches have effectively solved different kinds of FSSP, none of them considered the impact of manpower.
In practical manufacturing shops, the number of workers in the shop is limited, as well as their efficiency is diverse.
Therefore, besides the machines assignment, it is also imperative to assign these workers reasonably.
With the limitations of technology, a considerable part of the labor force still cannot be replaced by machines.
Since workers have diverse proficiency for stages, workers' arrangement is vital to production efficiency.

Some studies have considered the workers' influence during machine scheduling.
Marichelvam et al. \cite{ref19} presented an improved particle swarm optimization algorithm to solve HFSSP, while considering the various levels of the labor force and the effects of their learning and forgetting.
For the same problem, Behnamian \cite{ref28} proposed an improved colonial competitive algorithm by taking the effects of workers' learning and deterioration into consideration.
However, the above studies still focus on the machine arrangement, rather than the manpower scheduling.

There are some researches on the manpower scheduling for HFSSP, either without workers' proficiency or with high computational complexity.
Costa et al. \cite{ref21} proposed a novel discrete backtracking search algorithm powered by tabu search without considering the workers' proficiency for each stage.
Gong et al. \cite{ref30} proposed a hybrid evolutionary algorithm to solve the problem with worker flexibility, which means the spent time for each worker is different from manipulating each machine for each job.
However, both the above algorithms contain many dynamic parameters, which makes the poor performance with the high computational complexity or the limitation of local optimal.

In this paper, we concentrate on how to schedule manpower in the flow shop under the consideration of the workers' proficiency, which is defined as workers' production efficiency.
Our goal is to find the optimal workers' arrangement to minimize completion time when all products are processed.
Inspired by the Barnacle Mating Optimizer (BMO) algorithm with high computational efficiency, we adopted it to schedule the manpower.
However, BMO is only suitable for continuous space and can not be applied directly for our problem, which belongs to combinatorial optimization problems.
Therefore, combined Generic Algorithm (GA) with BMO, we proposed the Self-encoding Barnacles Mating Optimizer (SBMO) algorithm.
The main contributions of this paper are summarized as follows:
\begin{enumerate}
\item
For manpower scheduling, we define a Flow Shop Manpower Scheduling Problem (FSMSP) and establish a pure integer nonlinear programming model to solve this problem.
\item
Combined with the mutation and crossover operators, we extend the BMO algorithm to the combinatorial optimization problem.
Furthermore, we present the SBMO algorithm with a novel neighbor search scheme to minimize the completion time.
\item
Compared with the current popular algorithms, Improved Adaptive Genetic Algorithm (IAGA) and Muti-Objective Whale Swarm Algorithm (MOWSA), our proposed algorithm has higher performance with low computational complexity, stability, and a high approximate ratio.
\end{enumerate}

The rest of this paper is organized as follows.
Section \ref{sec2} briefly describes the problem and formulates the system model.
Next, Section \ref{sec3} illustrates the basic idea of the BMO algorithm.
Moreover, the SBMO algorithm is presented combined with GA.
In section \ref{sec4}, we conduct extensive simulations and analyze the performance results compared with current popular algorithms.
Finally, the last section presents the concluding remarks.

\section{Problem description and formulation}\label{sec2}
We define the FSMSP as follows.
The workers' proficiency is quantified to a decimal value between $(0,1]$, where the proficiency value is positively related to the workers' efficiency.
Specifically, there are the same $D$ products to be processed in $N$ stages, like clothes need to go through cutting, sewing, and ironing, etc..
The process order is fixed before production.
For simple description, we assume the process order is from 1 to $N$ in order.
In our shop model, there are $R$ workers to be allocated.
For convenience, we regard the spent time for a worker with a proficiency value of 1 to process a product as the \textit{unit time}, which of each stage is known.
It aims to arrange all workers for each stage to minimize the completion time $T$, which is defined as the time elapsed from the first product being processed until all $D$ products are finished.
For the production condition, it is required that each stage should arrange at least one worker.

From the perspective of \textit{Flow} concept, each stage is regarded as a network node, and each product is treated as a data packet to be processed.
Due to the short transmission distance between the two stages, the transfer time can be neglected.
That is, the product can be processed to the next idle stage without any interval.
In order to express the problem, the following notations are defined:

\textbf{Indices}

$i$: Index of workers, where $i$ = 1, 2, ... , $R$;

$j$: Index of stages, where $j$ = 1, 2, ... , $N$.

\textbf{Parameters}

$t_j$: The unit time of the $j^{th}$ stage;

$k_{ij}$: The proficiency of the $i^{th}$ worker to the $j^{th}$ stage;

$r_j$: The total number of workers arranged to the $j^{th}$ stage;

$T$: The completion time;

$D$: The number of products;

$X$: The workers' working status matrix;

$K$: The workers' proficiency matrix.

\textbf{Binary variable}

$$ x_{ij}=
\begin{cases}
1$, $\text{if the $i^{th}$ worker is assigned to the $j^{th}$ stage} \\
0$, $\text{otherwise}
\end{cases}$$

For convenience, we use the following two matrices to represent the workers' working status and their proficiency, where the row represents the worker, and the column represents the stage:

\begin{equation}
X=
\left[
\begin{array}{ccccccc}
	x_{11} & x_{12} & x_{13} & \cdots & x_{1j} & \cdots & x_{1N} \\
	x_{21} & x_{22} & x_{23} & \cdots & x_{2j} & \cdots & x_{2N} \\
	\vdots & \vdots & \vdots & \ddots & \vdots & \ddots & \vdots \\
	x_{i1} & x_{i2} & x_{i3} & \cdots & x_{ij} & \cdots & x_{iN} \\
	\vdots & \vdots & \vdots & \ddots & \vdots & \ddots & \vdots \\
	x_{R1} & x_{R2} & x_{R3} & \cdots & x_{Rj} & \cdots & x_{RN}
\end{array}
\right]
\end{equation}

\begin{equation}
K=
\left[
\begin{array}{ccccccc}
	k_{11} & k_{12} & k_{13} & \cdots & k_{1j} & \cdots & k_{1N} \\
	k_{21} & k_{22} & k_{23} & \cdots & k_{2j} & \cdots & k_{2N} \\
	\vdots & \vdots & \vdots & \ddots & \vdots & \ddots & \vdots \\
	k_{i1} & k_{i2} & k_{i3} & \cdots & k_{ij} & \cdots & k_{iN} \\
	\vdots & \vdots & \vdots & \ddots & \vdots & \ddots & \vdots \\
	k_{R1} & k_{R2} & k_{R3} & \cdots & k_{Rj} & \cdots & k_{RN}
\end{array}
\right]
\end{equation}
Take the value on the diagonal of $X^TK$ as the sum of the proficiency of all workers in the stage, that is: $\sum_{i=1}^{R}x_{ij}k_{ij}$.
Then the processing time required for a product in the $j^{th}$ stage is $t_j/\sum_{i=1}^{R}x_{ij}k_{ij}$.
We regard the processing time of each stage as its processing capacity, i.e., they have a negative relationship.

\begin{figure}[!t]
\centering
\includegraphics[width=3.7 in]{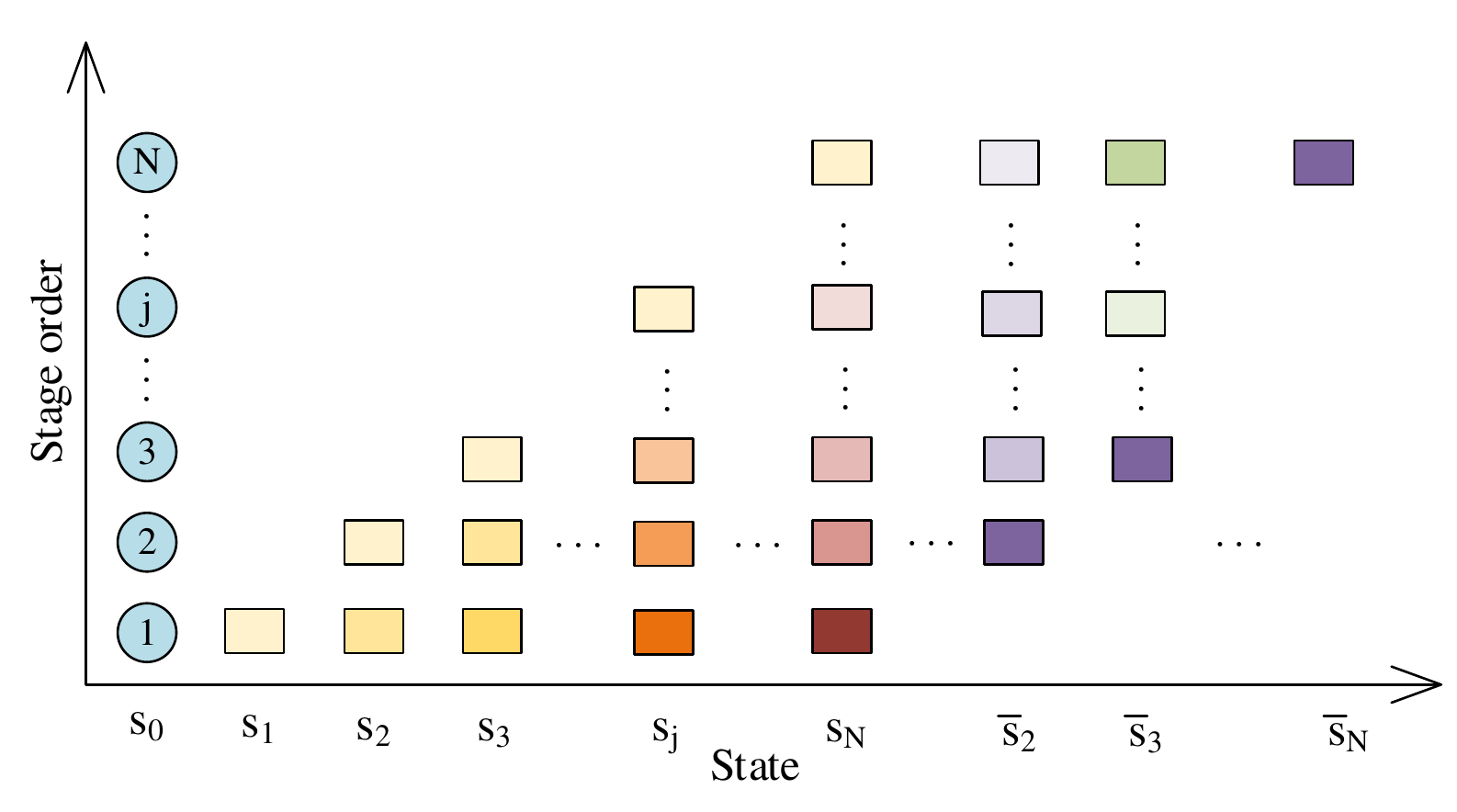}
\caption{Production states in the stages}
\label{order}
\end{figure}

The following describes how to calculate the completion time for the production of $D$ products.
As shown in the Fig. \ref{order}, there are $N$ stages and $2N-1$ production states.
The production states are defined as follows:

\textbf{Production states}

$s_0$: The initial state when products start to be processed;

$s_1$: The $1^{st}$ product complete the $1^{st}$ stage;

$s_2$: The $1^{st}$ product complete the $2^{nd}$ stage and the $2^{nd}$ product finish the $1^{st}$ stage;

$s_3$: The $1^{st}$ product complete the $3^{rd}$ stage, the $2^{nd}$ product finish the $2^{nd}$ stage and the $3^{rd}$ product finish the $1^{st}$ stage;

$s_j$: The $1^{st}$ product complete the $j^{th}$ stage, the $2^{nd}$ product finish the $({j-1})^{th}$ stage, and so on until the $j^{th}$ product finish the $1^{st}$ stage;

$s_N$: All $N$ stages have products to be processed. Note that this state occupies a period.

$\overline{s}_2$: After all products have completed the $1^{st}$ stage, the $D^{th}$ product complete the $2^{nd}$ stage, and the $({D-1})^{th}$ product finish the $3^{rd}$ stage, ..., the $({D-N+2})^{th}$ product finish the $N^{th}$ stage;

$\overline{s}_3$: After all products have completed the $2^{nd}$ stage, the $D^{th}$ product complete the $3^{rd}$ stage, and the $({D-1})^{th}$ product finish the $4^{th}$ stage, ..., the $({D-N+3})^{th}$ product finish the $N^{th}$ stage;

$\overline{s}_N$: After all products have completed the $({N-1})^{th}$ stage, the $D^{th}$ product complete the $N^{th}$ stage.
That is, all products finish their stages.

The processing time required for state $s_0$ to state $s_1$ is

\begin{equation}
\begin{split}
Y_1= \frac{t_1}{\sum_{i=1}^{R}x_{i1}k_{i1}}.
\end{split}
\end{equation}
After the first product is completed in the first stage, the time required for state $s_1$ to $s_2$ is

\begin{equation}
\begin{split}
Y_2=\max( \frac{t_1}{\sum_{i=1}^{R}x_{i1}k_{i1}}, \frac{t_2}{\sum_{i=1}^{R}x_{i2}k_{i2}}).
\end{split}
\end{equation}
By analogy, after $j^{th}$ stage is started, the time required for state $s_{j-1}$ to $s_j$ is

\begin{equation}
\begin{split}
Y_j=\max( \frac{t_1}{\sum_{i=1}^{R}x_{i1}k_{i1}}, \frac{t_2}{\sum_{i=1}^{R}x_{i2}k_{i2}},..., \frac{t_j}{\sum_{i=1}^{R}x_{ij}k_{ij}}).
\end{split}
\end{equation}

Since the first product reaches the last stage, state $s_N$ begins and lasts for a period of

\begin{equation}
\begin{split}
(D-N+1)Y_N=(D-N+1)\max( \frac{t_1}{\sum_{i=1}^{R}x_{i1}k_{i1}}, \\
\frac{t_2}{\sum_{i=1}^{R}x_{i2}k_{i2}},...,\frac{t_N}{\sum_{i=1}^{R}x_{iN}k_{iN}}).
\end{split}
\end{equation}
Next, all products have gone through the first stage.
At this time, the time required for state $s_N$ to $\overline{s_2}$ is

\begin{equation}
\begin{split}
\overline{Y}_2=\max( \frac{t_2}{\sum_{i=1}^{R}x_{i2}k_{i2}}, \frac{t_3}{\sum_{i=1}^{R}x_{i3}k_{i3}},...,\frac{t_N}{\sum_{i=1}^{R}x_{iN}k_{iN}}).
\end{split}
\end{equation}
When all products have gone through the second stage, the time required for state $\overline{s}_2$ to $\overline{s}_3$ is

\begin{equation}
\begin{split}
\overline{Y}_3=\max( \frac{t_3}{\sum_{i=1}^{R}x_{i3}k_{i3}}, \frac{t_4}{\sum_{i=1}^{R}x_{i4}k_{i4}},..., \frac{t_N}{\sum_{i=1}^{R}x_{iN}k_{iN}}).
\end{split}
\end{equation}
By parity of reasoning, after all products have completed the ${(j-1)}^{th}$ stage, the time required for state $\overline{s}_{j-1}$ to $\overline{s}_j$ is

\begin{equation}
\begin{split}
\overline{Y}_j=\max( \frac{t_j}{\sum_{i=1}^{R}x_{ij}k_{ij}}, \frac{t_{j+1}}{\sum_{i=1}^{R}x_{i(j+1)}k_{i(j+1)}},..., \\
\frac{t_N}{\sum_{i=1}^{R}x_{iN}k_{iN}}).
\end{split}
\end{equation}
Finally, the last product is processed in the last stage, the required time is

\begin{equation}
\begin{split}
\overline{Y}_N= \frac{t_N}{\sum_{i=1}^{R}x_{iN}k_{iN}}.
\end{split}
\end{equation}
Then, the total completion time is:

\begin{equation}
\begin{split}
T=Y_1+Y_2+\cdots+Y_j+\cdots+(D-N+1)Y_N+\\
\overline{Y}_2+\overline{Y}_3+\cdots+\overline{Y}_j+\cdots+\overline{Y}_N\label{makespan}.
\end{split}
\end{equation}

With the notations above, we present the following Pure Integer Nonlinear Programming (PINLP) model for the problem:

\begin{equation}
\begin{split}
\min T=Y_1+Y_2+\cdots+Y_j+\cdots+(D-N+1)Y_N\\
+\overline{Y}_2+\overline{Y}_3+\cdots+\overline{Y}_j+\cdots+\overline{Y}_N\label{makespan2}
\end{split}
\end{equation}


\begin{equation}
\sum_{j=1}^{N}x_{ij}=1\label{cons1}
\end{equation}

\begin{equation}
r_j \geq1\label{cons2} \ with \ r_j\in Z
\end{equation}

\begin{equation}
\sum_{j=1}^{N}r_j=R\label{cons3}
\end{equation}

\begin{equation}
t_j\geq0\label{cons4}
\end{equation}


\begin{equation}
D\gg N\label{cons6}
\end{equation}

\begin{equation}
R\geq N\label{cons7}
\end{equation}

As shown in the Eqs. (\ref{makespan2}), our optimization goal is to minimize the completion time.
Constraint (\ref{cons1}) requires that each worker can only be assigned to one stage;
constraint (\ref{cons2}) ensures that each stage should arrange at least one worker and the number of workers arranged in each stage must be an integer, where $r_j=\sum_{i=1}^{R}x_{ij}$;
constraint (\ref{cons3}) guarantees the total number of workers to be $R$;
constraint (\ref{cons4}) restricts the unit time of each stage is greater than 0;
constraint (\ref{cons6}) requires that the number of products is much greater than the number of stages, which is feasible in a real industrial shop.
For example, the amount of most essential stages in garment production is only four, including designing/clothing pattern generation, fabric cutting, sewing, and ironing/packing \cite{ref31}, while the number of normal orders of clothing companies is at least two hundred \cite{ref32}.
Constraint (\ref{cons7}) requires that the number of workers is not smaller than the number of stages to make sure each stage has at least one worker.

Before digging into any algorithms, we first analyze the solution space of the problem.
The number of possible solutions for assigning $R$ workers to $N$ stages is $C_{R-1}^{N-1}*A_N^N$.
Such an ample solution space requires the effective utilization of problem-specific properties to design efficient algorithms.

\section{Self-encoding barnacles mating algorithm}\label{sec3}

In this section, we first give the basic idea of the BMO algorithm.
In order to make the BMO algorithm be applied for the FSMSP, we illustrate the combination of GA and BMO algorithms.
To solve the problem of local optimum, we design a neighborhood search method compromising mating behavior.
Furthermore, we present the SBMO algorithm to handle the FSMSP and finally give the analysis of SBMO complexity.

\subsection{The basic idea of BMO algorithm}
BMO was first developed by Mohd Herwan Sulaiman et al. in 2018\cite{ref1}, which is inspired by the mating behavior of real barnacles.
Since the barnacle is hermaphrodite, it can provide its sperm and accept sperm from other barnacles.
To cope with tidal changes and a sedentary lifestyle, the penis of a barnacle can reach 7 to 8 times its body length, and all the neighbors that the penis can reach constitute its mating population.
The BMO algorithm can be divided into the following steps.

\subsubsection{Solution mode}
In BMO, it is supposed that the candidate solutions are the barnacles where the vector of the population can be expressed as follows:
\begin{equation}
V=
\left[
\begin{array}{ccccc}\label{BMO}
v_1^1 & \cdots & v_1^j & \cdots & v_1^W\\
\vdots & \ddots & \vdots & \ddots & \vdots\\
v_i^1 & \cdots & v_i^j & \cdots & v_i^W\\
\vdots & \ddots & \vdots &\ddots & \vdots\\
v_w^1 & \cdots & v_w^j & \cdots & v_w^W
\end{array}
\right]
\end{equation}
where $v_i^j$ is the control variable of the solution.
$W$ is the number of control variables, and $w$ is the population size, that is, the number of barnacles in the population.
Each row vector has a corresponding function value to represent the performance of this solution.
To retain the elites, row vectors of $V$ are sorted by their function values.

\subsubsection{Offspring generation}\label{generation}
The traditional evolutionary algorithms often use complex selection methods, while the BMO algorithm only exploits the simple random selection for two barnacles: the father and the mother.
There are two mating behaviors in BMO, including mating and sperm-cast mating.
Specifically, the mating behavior is determined by the relationship between the penis length ($pl$) and barnacles’ distance.
In detail, if the distance is smaller than $pl$, then the selected two barnacles mate; otherwise, the sperm-cast mate will occur.

\paragraph{Mating}
As shown in Eqs. (\ref{mate}), the mating rule is to inherit the father’s characteristics with probability $p$ and the ones from the mother with probability $1-p$.
\begin{equation}
v_i^{W\_new}=pv_{barnacle\_f}^W+(1-p)v_{barnacle\_m}^W, \label{mate}
\end{equation}
where $v_{barnacle\_f}^W$ and $v_{barnacle\_m}^W$ are the variables of the vectors from the father and mother, respectively.

\paragraph{Sperm-cast mating}
As shown in Eqs. (\ref{m-mate}), sperm-cast mating is the mother’s self-mutation.
\begin{equation}
v_i^{W\_new}=rand() \times v_{barnacle\_m}^W \label{m-mate}
\end{equation}
where $rand()$ is the random number within [0,1].

\subsubsection{Population update rules}
After the offspring generation is finished, the new offspring are merged with their parents.
In order to keep population size stable, the individuals corresponding poor solutions will be filtered.
The population evolution continues until its stopping condition is met.

BMO has the advantages of few parameters and fast convergence speed.
However, the control variables in the BMO algorithm search in continuous space, while the control variables of FSMSP (the workers’ schedule) are the binary matrix.
Hence, FSMSP belongs to the combinatorial optimization problem, and the BMO algorithm can not be applied directly.
GA is a popular tool to deal with the combinatorial optimization problem, but normally has amounts of parameters and is easy to be trapped into the local optimum.
It is tended to cooperate with the GA and BMO metaheuristic algorithms to solve FSMSP.

\subsection{Combination GA and BMO}\label{sec:combination}
GA is a random global search algorithm, which encodes the discrete variables and iterates generations by selection, crossover, and mutation to obtain the optimal solution.
Specifically, we modify the solution mode and the offspring generation schemes (Eqs. (\ref{mate}) and Eqs. (\ref{m-mate})) in BMO with encoding, crossover, and mutation in GA.
The highlight of this combination is to encode the solutions of the FSMSP, as well as design crossover and mutation operators.


\subsubsection{Solution mode}
\begin{figure}[!t]
\centering
\includegraphics[width=3.0 in]{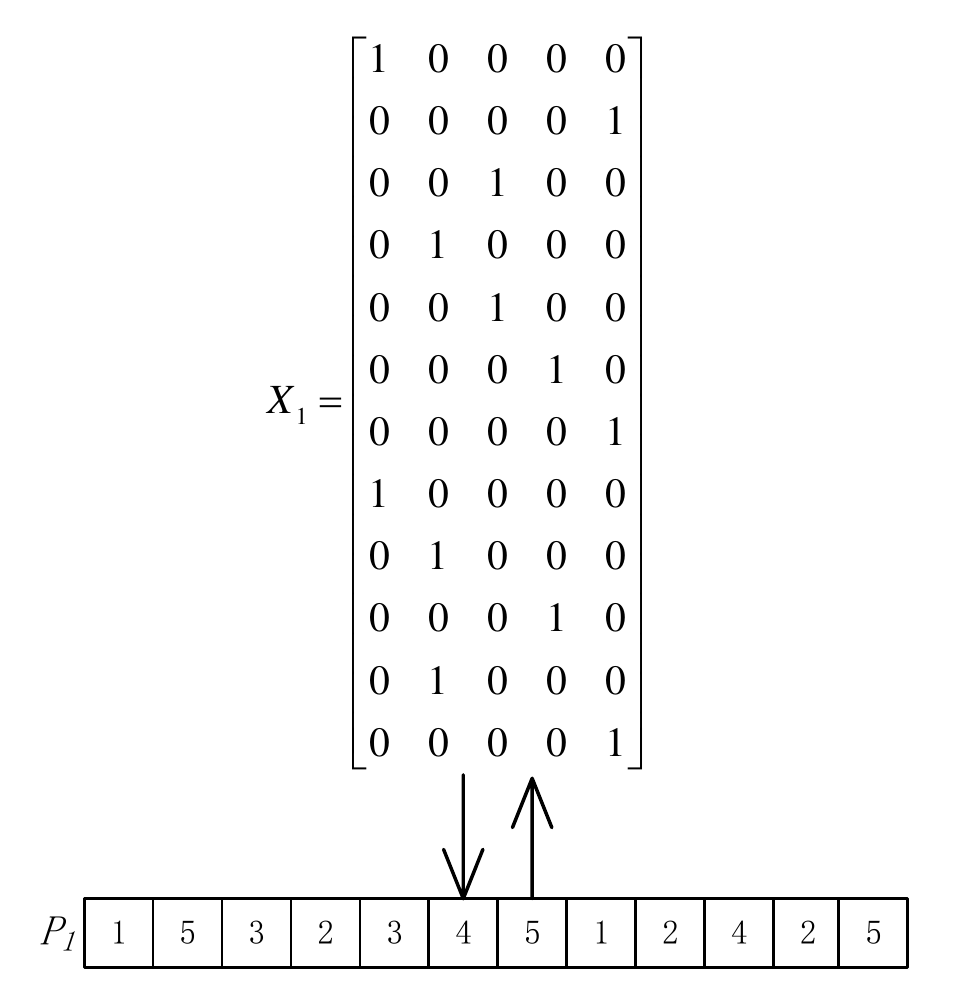}
\caption{Transformation of the matrix by assigning 12 workers to 5 stages}
\label{string}
\end{figure}

The solution mode of FSMSP is a matrix, while the one of BMO is a row vector.
Thus, it is necessary to transfer the solution mode from the matrix to a row vector.
The number of row vectors is recorded as the corresponding positions of 1 for each line of matrix $X$.
Hence, the length of the row vector is the total number of allocated workers.
The $i^{th}$ number $j$ in the row vector represents that the $i^{th}$ worker is assigned to the $j^{th}$ stage.
Fig. \ref{string} illustrates the transformation for a specific matrix.
Matrix $X_1$ is one of the FSMSP solutions, indicating a worker schedule by assigning 12 workers to 5 stages.
For the transferred solution mode $P_1$, the third number 3 means that the third worker is arranged to the third stage in the worker schedule.
It is mentioned that the decoding method used in the following computation of the completion time is the inverse operation above.

\subsubsection{Offspring generation}
In this part, we replace the mating and sperm-cast mating methods of BMO with the crossover and mutation operators of GA.

\begin{figure}[!t]
\centering
\includegraphics[width=3.0 in]{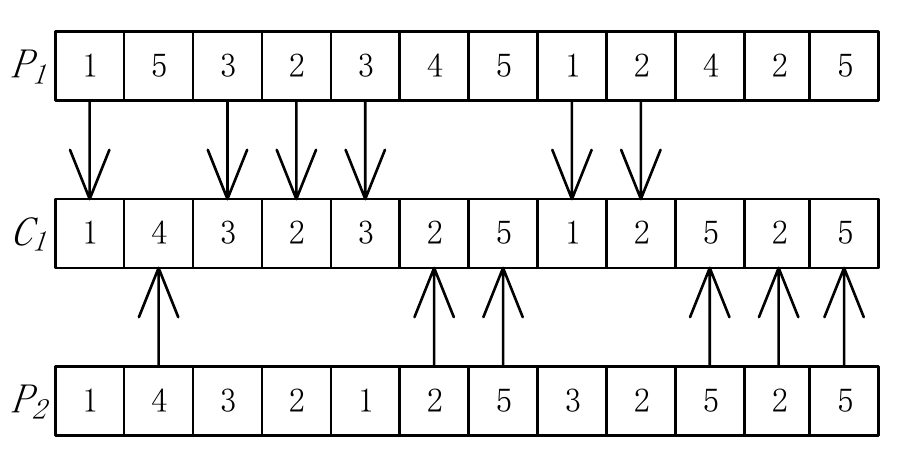}
\caption{The illustration of crossover operator}
\label{crossover}
\end{figure}

\paragraph{Crossover operator}
The crossover operator is modified from the mating behavior, which is designed to copy the row vectors of the mother and the father, both with 50\% probability.
As shown in Fig. \ref{crossover}, the row vector $P_1$ and $P_2$ cross to produce offspring $C_1$.
The up-to-down arrow means the variable of $C_1$ copies from the one of $P_1$.
On the contrary, the down-to-up arrow notes the variable of $C_1$ copies from the one of $P_2$.
Each crossover only generates one offspring.

\begin{figure}[!t]
\centering
\includegraphics[width=3.0 in]{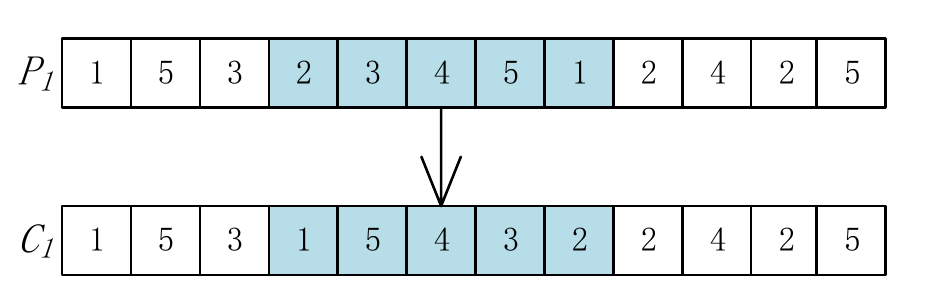}
\caption{The illustration of the inversion mutation}
\label{mutation1}
\end{figure}

\begin{figure}[!t]
\centering
\includegraphics[width=3.0 in]{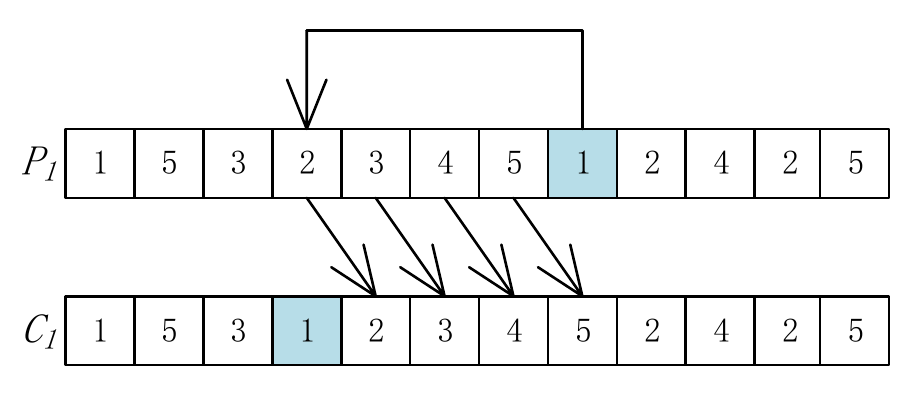}
\caption{The illustration of the insertion mutation}
\label{mutation2}
\end{figure}

\begin{figure}[!t]
\centering
\includegraphics[width=3.0 in]{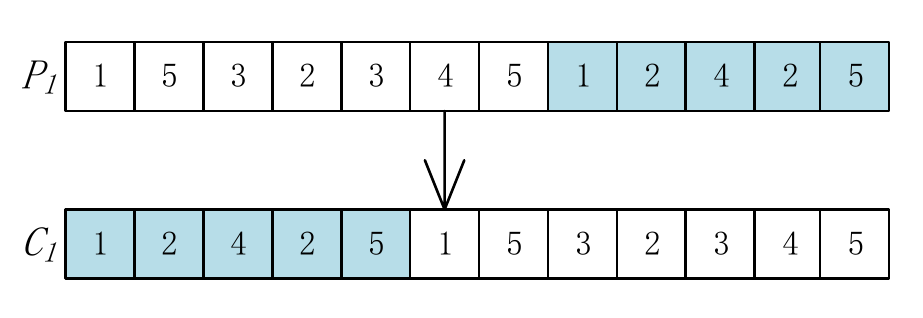}
\caption{The illustration of the double-segment swap mutation}
\label{mutation3}
\end{figure}

\paragraph{Mutation operator}
This operator aims to modify the sperm-cast mating behavior.
To make the best of population diversity, we adopt three mutation methods, including inversion mutation \cite{ref24}, insertion mutation \cite{ref25}, and double-segment swap mutation.
When the mother barnacle undergoes self-mutation, it will choose one of the above randomly.
\begin{itemize}
\item Inversion mutation ($M_1$):
Two positions are selected randomly within the row vector $P_1$.
Reverse the sequence of the variables between two positions to generate the offspring $C_1$, illustrated as Fig. \ref{mutation1}.
\item Insertion mutation ($M_2$):
A variable on $P_1$ is randomly selected and moved to other positions to generate offspring $C_1$, as shown in Fig. \ref{mutation2}.
\item Double-segment swap mutation ($M_3$):
$P_1$ is randomly divided into two segments, which are exchanged to generate offspring $C_1$, seen from Fig. \ref{mutation3}.
\end{itemize}

\begin{figure}[!t]
\centering
\includegraphics[width=3.1 in]{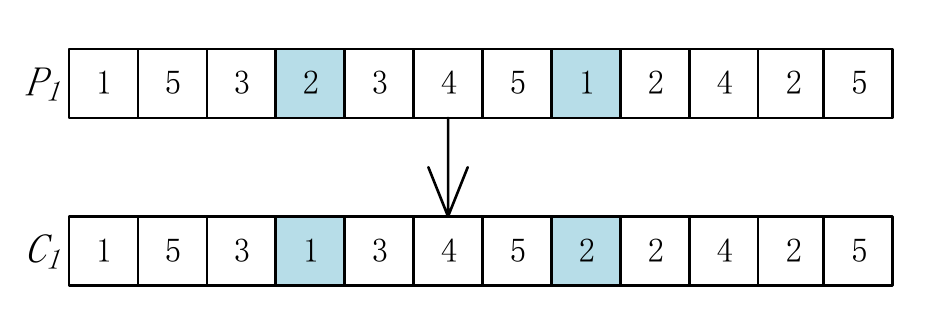}
\caption{The illustration of the reciprocal exchange mutation}
\label{neighbour1}
\end{figure}

\begin{figure}[!t]
\centering
\includegraphics[width=3.1 in]{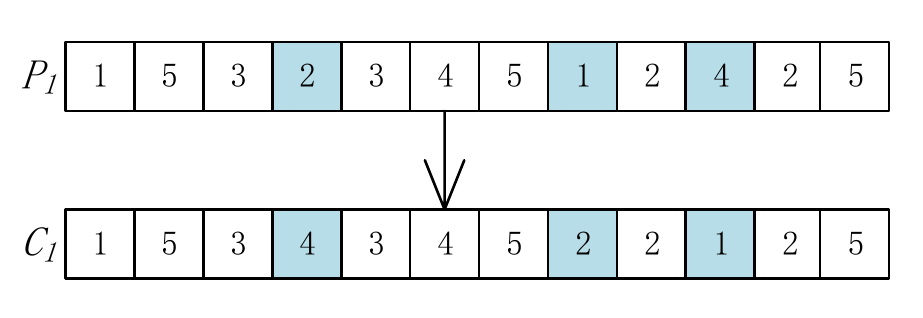}
\caption{The illustration of the neighborhood mutation}
\label{neighbour2}
\end{figure}

\subsubsection{Population update}
It is retained the population innovation rules in the BMO algorithm.
It is noted that the function value of each row vector is defined as the completion time.

\subsection{Neighborhood search}
\label{sec:neighbor}
Along with the generation iterations, the updated population tends to the optimal solution.
Hence, the filtered solutions tend to have high similarities.
The crossover operation fails to generate new offsprings, which makes the solution locked into a local optimum.
To solve the above problem, we propose a neighborhood search algorithm to improve the offspring generation.

\textbf{Offspring generation:}
When the feasible solutions tend to be the same, there is no meaning to do the crossover operation.
Therefore, we set $pl$ as 0 to block the crossover.
Meanwhile, we adopt two mutation operators and design a novel one to fine-tune solutions.
We randomly choose one mutation operator to generate offspring.

\paragraph{Crossover operator}
This operator is blocked by setting $pl$ as 0.

\paragraph{Mutation operator}
According to the Cannikin law \cite{ref33}, the water capacity of the wooden barrel is determined by the shortest wooden board.
Thus, the whole process capacity for the FSMSP is constrained by the weakest stage capacity.
In order to make stage capacity balanced, we put forward a novel mutation operator called \emph{balance mutation ($M_4$)}.
The main idea is to randomly pick up one worker from the stage with the most robust capacity for the weakest one.

However, the unique mutation operator fails to guarantee population diversity.
Therefore, we expend two other mutation operators: reciprocal exchange mutation ($M_5$) \cite{ref26} and triplet mutation ($M_6$) \cite{ref25}.
At the reciprocal exchange mutation, two variables are selected randomly and swapped, as shown in Fig. \ref{neighbour1}.
At the triplet mutation, it is random to select three variables and swap, demonstrated in Fig. \ref{neighbour2}.

\subsection{The SBMO algorithm}
Combined with BMO, GA, and neighborhood search scheme, we propose the SBMO algorithm to solve FSMSP, outlined in Algorithm \ref{alg1}.

\subsubsection{Solution mode (Lines 1-6)}
The algorithm first generates $Q$ legitimate 0-1 matrices $X$ with only one 1 in each row.
Next, it calculates the function value of each $X$ and encodes them into corresponding row vectors (barnacles), which constitute the original population.
Then sort the barnacles from superior to inferior according to their function values.

\subsubsection{Offspring generation (Lines 8-16)}
In Algorithm \ref{alg1}, $Dis$ is the distance between the father and the mother.
If $Dis$ is within $pl$, the two barnacles can mate, i.e., the two row vectors cross.
Otherwise, sperm-cast mating occurs, i.e., the mother self-mutation.
The mutation operator is randomly from $M_1$ to $M_3$, illustrated in the previous section \ref{sec:combination}.

\subsubsection{Population update (Lines 17-22)}
After merging the legal offsprings and the parents, each barnacle must be decoded into a matrix $X$ for easy computation of function values.
Sort the barnacles from superior to inferior by function values.
In order to keep the population scale constant, it is retained the $Q$ superior barnacles.

\subsubsection{Neighborhood search (Lines 23-25)}
When all barnacles in the population are the same, $pl$ is set to 0.
Then the neighborhood search algorithm is triggered, where the mutation operator is chosen randomly from $M_4$ to $M_6$, described in previous section \ref{sec:neighbor}.
The generation will be iterated for $G$ rounds to obtain the final solution.

\begin{algorithm}[H]
\renewcommand{\algorithmicrequire}{\textbf{Input:}}
\renewcommand{\algorithmicensure}{\textbf{Output:}}
\caption{SBMO}
\label{alg1}
\begin{algorithmic}[1]
\Require Workers' proficiency $K$, the number of products $D$, the unit time $t_j$, the number of maximum generations $G$, the population size $Q$ and $pl \in [0,Q]$;
\Ensure Worker schedule $X$;
\State Initialize: Generate $Q$ legitimate matrices $X$ randomly.
\For{ $q=1$ to $Q$}
\State Calculate the function value $T_q$ of each $X$.
\EndFor
\State Encode all $X$ into corresponding row vectors (barnacles);
\State Sort the barnacles from superior to inferior by $T_q$;
\For{$g=1$ to $G$}
\For{$q=1$ to $Q$}
\State Randomly select one father ($barnacle\_f$) and one mother ($barnacle\_m$);
\State $Dis=dis(barnacle\_f, barnacle\_m)$;
\If{$Dis \leq pl$}
\State Two barnacles mating (crossover operators);
\Else
\State Sperm-cast mating (one mutation operator is chosen randomly from $M_1$ to $M_3$);
\EndIf
\EndFor
\State Delete the illegal offsprings by the constraint conditions;
\State Merging offsprings and parents;
\State Decode each barnacle into matrix $X$;
\State Calculate the function value $T_q$ of all $X$;
\State Sort the barnacles from superior to inferior by $T_q$;
\State Retain the $Q$ superior barnacles;
\If{All solutions in the population are the same}
\State Set $pl$ to 0;
\State One mutation operator is chosen randomly from $M_4$ to $M_6$;
\Else
\State Continue;
\EndIf
\EndFor
\end{algorithmic}
\end{algorithm}

\subsection{Algorithm complexity}
In the initialization phase, we adopt the random method to generate the legitimate matrices $X$.
 We first generate an $N*N$ identity matrix and another random 0-1 matrix with $(R-N)*N$ size where each row has only one 1 and zeros elsewhere. 
Next, we combine the above matrices to generate a new matrix with $R*N$ size, and randomly scramble the positions of rows to generate matrix $X$.
The above method can guarantee that the matrix $X$ satisfy the constraints of Eqs. \ref{cons1}-\ref{cons3}.
The time complexity of \textit{initialization} is $O(QR)$.

According to Eqs. (\ref{makespan}), the time complexity of calculating the function value of each $X$ (Line 3) is $O(2N-2)=O(N)$.
The \textit{for-loop} runs for $Q$ times.
Hence, the \textit{for-loop} runs in $O(QN)$.
The encode from each $X$ to a row vector costs $O(R)$, and there are $Q$ matrices $X$ to be encoded.
Line 5 takes $O(QR)$.
The sorting of $Q$ numbers (Line 6) consumes $O(Q\log Q)$.
In practice, the population size $Q$ usually chooses below 1000, and the number of workers $R$ is often more than 10.
Since $\log Q<10$ and $R>10$, the time complexity of \textit{solution mode} is $O(QR)$.

In the \textit{for-loop} (Lines 8-16), the distance computation for two barnacles (Lines 9-10) costs constant time.
For the crossover operator, the offspring generation copies variables from the father or the mother one by one.
The number of variables for the offspring is the number of workers $R$.
Hence the crossover operator (Line 12) takes $O(R)$.
The mutation operator is randomly from $M_1$ to $M_3$.
It is easy to obtain that the time complexity of $M_1$, $M_2$ and $M_3$ are constant, $O(R-1)$ and $O(R)$, respectively.
Thus, the mutation operator (Line 14) takes $O(R)$ too.
Therefore, the time complexity of \textit{offspring generation} is $O(QR)$.

For Line 17, the operation to judge the illegal offspring takes $O(QN)$.
The merge operation consumes constant time (Line 18).
The decoding operation is opposite to the encoding one, and the time complexity is the same, i.e., $O(QR)$.
The operation of Line 20 is similar to Lines 2-4, and it takes $O(QN)$.
Line 21 is the same as Line 6, and it runs in $O(2Q\log ({2Q}))=O(Q\log Q)$.
Line 22 takes constant time. Hence, the time complexity of \textit{population update} is $O(QR)$.

It is evident that Lines 23-24 take the constant time.
The time complexity of each mutation operator from $M_4$ to $M_6$ is constant, too.
Hence, the time complexity of \textit{neighborhood search} is constant.

In summary, the whole time complexity of the proposed SBMO algorithm is $O(GQR)$.

\section{Experimental results}\label{sec4} 
To evaluate the performance of the SBMO, we exploit the following metrics through simulation experiments.
\begin{itemize}
\item [1.]
Completion time ($T$): The time is calculated by Eqs. (\ref{makespan}) according to the worker schedule $X$.
\item [2.]
Approximation ratio ($\gamma$): This is the main metric demonstrating the performance of the SBMO algorithm.
It demonstrates how the SBMO algorithm approaches the globally optimal solution ($T^*$), calculated by Lingo in our simulation.
It is noted that $T^*$ can be obtained only in small scale.
\begin{equation}
\gamma=\frac{T}{T^*} \label{AR},
\end{equation}
where $T$ is the obtained completion time by using various algorithms.
\item [3.]
Standard Deviations ($SD$): Given the fixed scale, it is used to evaluate the stability of obtained solutions.
\item [4.]
Execution time: The total time of algorithm execution is the time cost to find optimal solutions.
\end{itemize}


\begin{table}
\centering
\footnotesize
\renewcommand\tabcolsep{2.5pt} 
\caption{Comparison results for the proposed neighborhood search algorithm }
\begin{tabular}{ccccccc} 
\toprule
\multirow{2}{*}{Problem} & \multicolumn{2}{c}{$\gamma$} & \multicolumn{2}{c}{$SD$} & \multicolumn{2}{c}{Execution time(s)}  \\
\cmidrule(r){2-3}  \cmidrule(r){4-5} \cmidrule(r){6-7}

   & SBMO & SBMO-WN & SBMO & SBMO-WN & SBMO & SBMO-WN \\
  \midrule
  4-12 & \textbf{1.0000} & \textbf{1.0000}  & \textbf{0.00} & \textbf{0.00}  & 25.25 & 19.97 \\
  4-16 & 1.0013 & 1.0017  & 0.85 & 0.89  & 28.08 & 21.05 \\
  4-20 & \textbf{1.0000} & 1.0008  & \textbf{0.00} & 0.33  & 30.20 & 24.08 \\
  4-24 & \textbf{1.0000} & \textbf{1.0000}  & \textbf{0.00} & \textbf{0.00}  & 31.63 & 26.30 \\
  4-28 & \textbf{1.0000} & \textbf{1.0000}  & \textbf{0.00} & \textbf{0.00}  & 31.25 & 27.78 \\
  4-32 & \textbf{1.0000} & \textbf{1.0000}  & \textbf{0.00} & \textbf{0.00}  & 32.07 & 27.57 \\
  6-12 & \textbf{1.0000} & \textbf{1.0000} & \textbf{0.00} & \textbf{0.00} & 25.12 & 18.28\\
  6-16 & 1.0120 & 1.0141  & 7.10 & 6.43 & 30.33 & 23.72\\
  6-20 & 1.0062 & 1.0158  & 3.14 & 5.24 & 29.65 & 24.11 \\
  6-24 & 1.0016 & 1.0082  & 0.96 & 2.81 & 31.00 & 26.89 \\
  6-28 & 1.0003 & 1.0150  & 0.24 & 3.27 & 33.41 & 28.93 \\
  6-32 & 1.0009 & 1.0120  & 0.65 & 1.49 & 34.31 & 27.39 \\
  8-12 & \textbf{1.0000} & 1.0003  & 0.20 & 0.26  & 26.51 & 24.50 \\
  8-16 & 1.0047 & 1.0097  & 11.00 & 14.62  & 26.79 & 26.97 \\
  8-20 & 1.0094 & 1.0221  & 3.53 & 4.91 & 30.38 & 27.63 \\
  8-24 & 1.0014 & 1.0057  & 3.10 & 3.28 & 30.11 & 29.84 \\
  8-28 & 1.0095 & 1.0198  & 7.46 & 5.50 & 32.16 & 30.78 \\
  8-32 & 1.0002 & 1.0103  & 0.22 & 3.69 & 36.88 & 37.45 \\
  10-12 & \textbf{1.0000} & 1.0028  & \textbf{0.00} & 5.53  & 26.52 & 22.39 \\
  10-16 & 1.0019 & 1.0076  & 5.05 & 6.50  & 30.78 & 23.40 \\
  10-20 & 1.0037 & 1.0250  & 13.58 & 16.83 & 30.50 & 29.85 \\
  10-24 & 1.0066 & 1.0095  & 5.63 & 4.94 & 31.57 & 29.71 \\
  10-28 & 1.0002 & 1.0073  & 0.85 & 5.53 & 36.24 & 36.31 \\
  10-32 & 1.0092 & 1.0316  & 3.85 & 6.74 & 40.08 & 39.51 \\
  12-12 & 1.0003 & 1.0061  & 1.26 & 194.57 & 20.03 & 18.93 \\
  12-16 & 1.0030 & 1.0357  & 17.03 & 49.26 & 26.74 & 27.28 \\
  12-20 & 1.0122 & 1.0877  & 29.79 & 27.97 & 31.15 & 26.33 \\
  12-24 & 1.0081 & 1.0481  &7.69 & 25.76 & 32.13 & 33.02 \\
  12-28 & 1.0147 & 1.0445  & 12.84 & 10.98 & 38.00 & 32.48 \\
  12-32 & 1.0048 & 1.0174  & 3.23 & 8.03 & 39.23 & 39.30 \\
  \bottomrule
  \end{tabular}
  \label{table1}
\end{table}

\subsection{Simulation setup}
All simulations ran on a PC with an Intel Core i7 1.8-GHz and 8 GB memory.
Each simulation is repeated 20 times, and the average values are recorded as statistical results.

We adopt the IAGA \cite{ref3} and MOWSA \cite{ref2} as the comparisons.
The IAGA is a popular tool to address the combinational optimization problem.
The MOWSA was designed to solve the distributed permutation flow shop scheduling problem, which can be applied for FSMSP by modifying the encoding objective from jobs to workers.
Since the neighbor search scheme presented in the MOWSA focuses on the schedule of jobs and factories, the scenario is not suitable for FSMSP.
We only implement the MOWSA without its neighbor search part.
For the sake of fairness, all algorithms use the same crossover operator and the mutation operators selection rule(chosen randomly from $M_1$ to $M_3$).

\begin{figure}[!t]
\centering
\includegraphics[width=3.5 in]{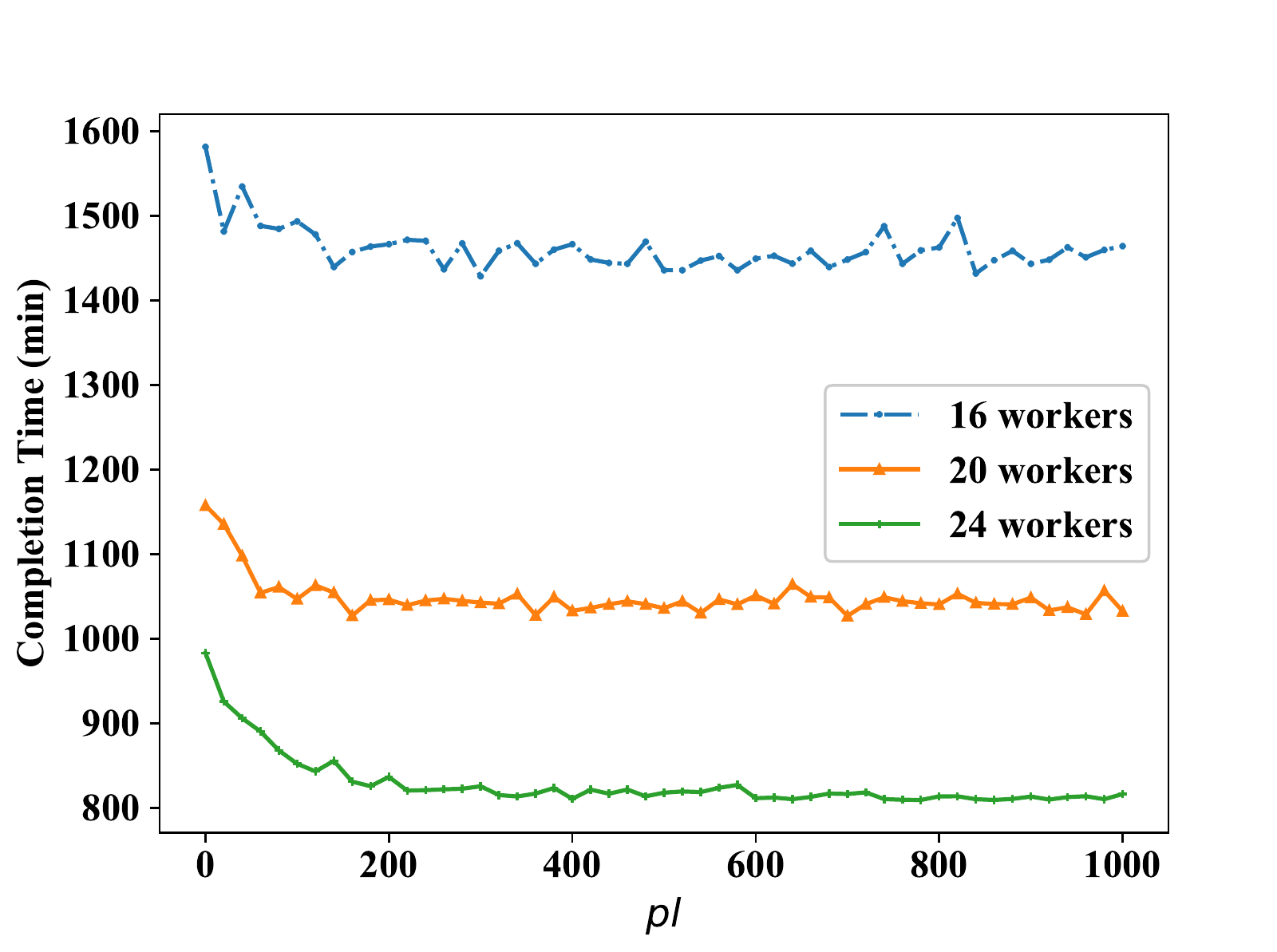}
\caption{$pl$ determination with 12 stages}
\label{pl}
\end{figure}

The proposed algorithm needs to define three parameters: population size $Q$, the number of maximum generations $G$, and $pl$.
From the extensive simulation results, it is induced that $Q=1000$ and $G=500$ are the feasible settings to obtain the optimal solutions.
All the other parameters involved in compared algorithms are referred to the settings in their literature.
To determine the $pl$ setting, we evaluate $T$ with the increasing $pl$ from 0 to 1000, in three cases of the various number of workers (16, 20, and 24 workers) and 12 stages.
As can be seen from Fig. \ref{pl}, before $pl$ reaches 200, $T$ decreases obviously with the rising $pl$.
While $pl$ is greater than 200, $T$ tends to converge.
The reason is that the diversity of the offspring raises with the increasing $pl$ at the first period ($pl \in [0, 200]$), which brings in the higher probability of the superior solutions.
When $pl > 200$, the crossover operator occupies the majority, which results in the solution diversity on the edge of disappearance.
At this time, the neighborhood search scheme is executed to improve the solution, which can usually obtain the approximate optimal solution.
Hence, $T$ vibrates slightly near the optimal solution.

Fig. \ref{pl} further reveals that the fluctuation of $T$ relieves the expanding number of workers.
With the larger amount of workers, the adjustment of workers has less impact on each stage capacity, with the same effect on $T$.
Therefore, we random set the $pl$ among [200, 1000) in our simulations.

\begin{figure*}[htbp]
\centering  
\subfigure[The case of 20 workers and 12 stages]
{
\includegraphics[width=3.4 in]{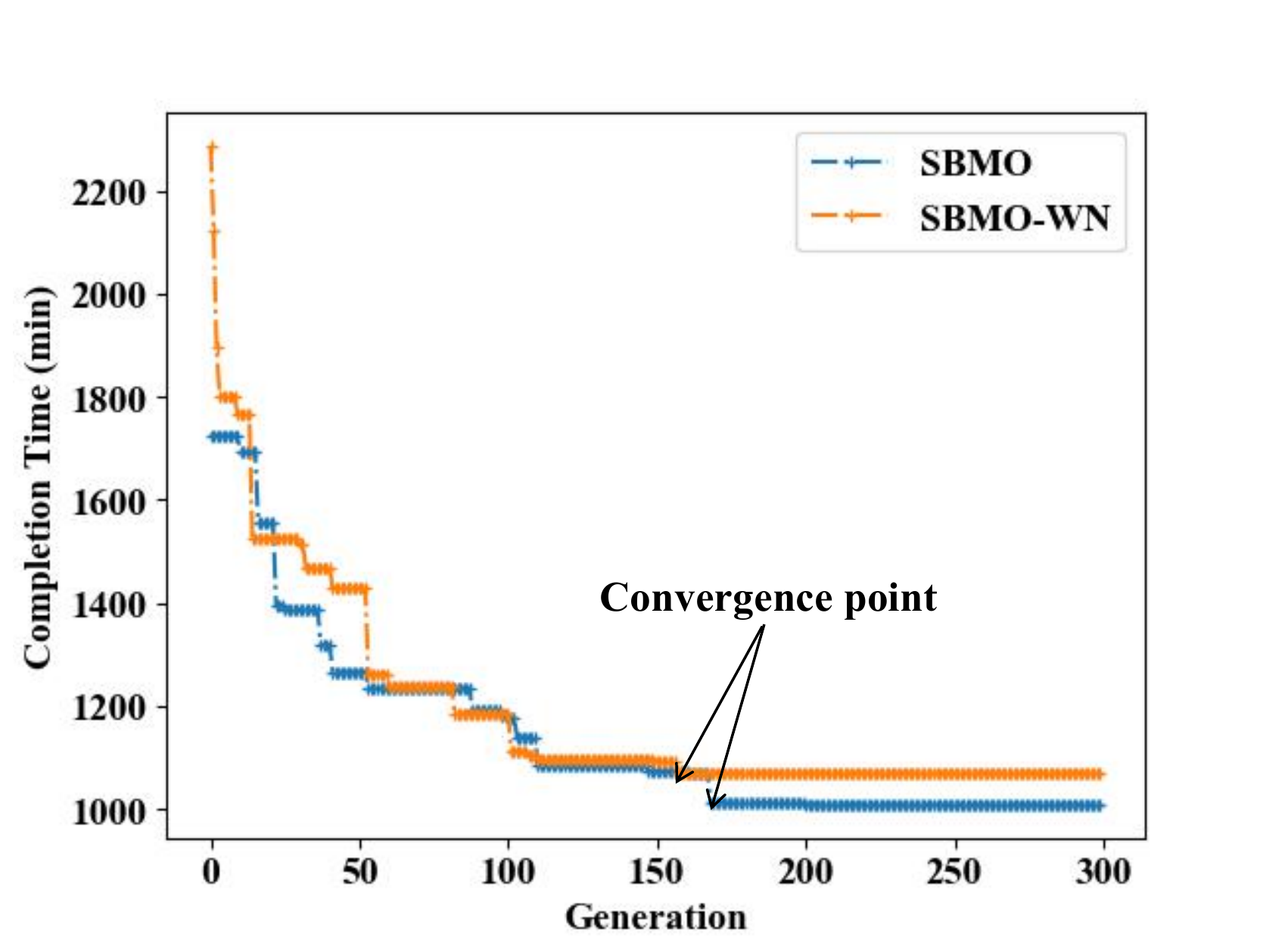}
}
\subfigure[The case of 30 workers and 12 stages]
{
\includegraphics[width=3.4 in]{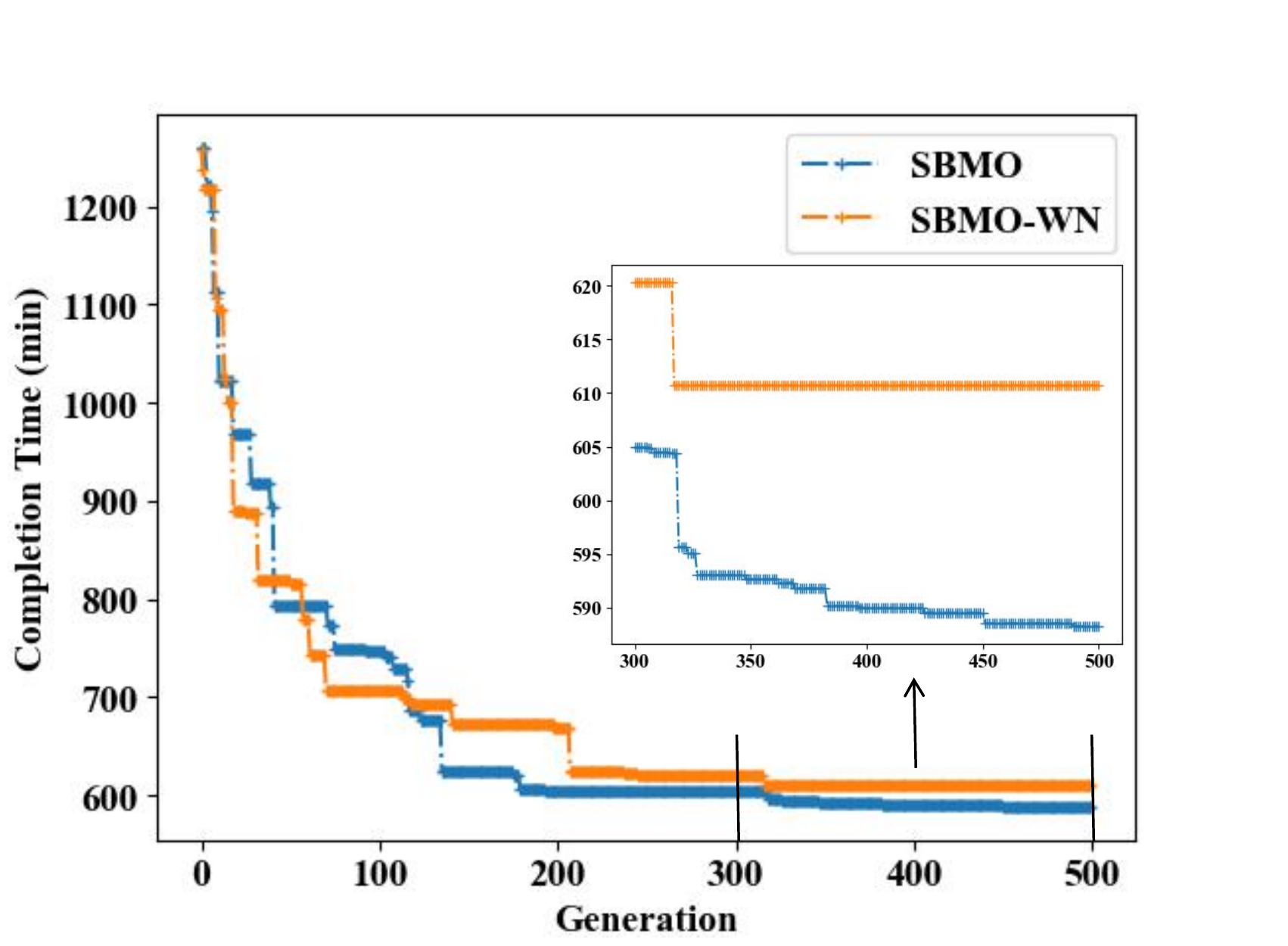}  
}
\caption{Convergence curves}
\label{generation1}
\end{figure*}

\begin{figure*}[htbp]
\centering

\subfigure[4-stage cases]{
\includegraphics[width=2.22 in]{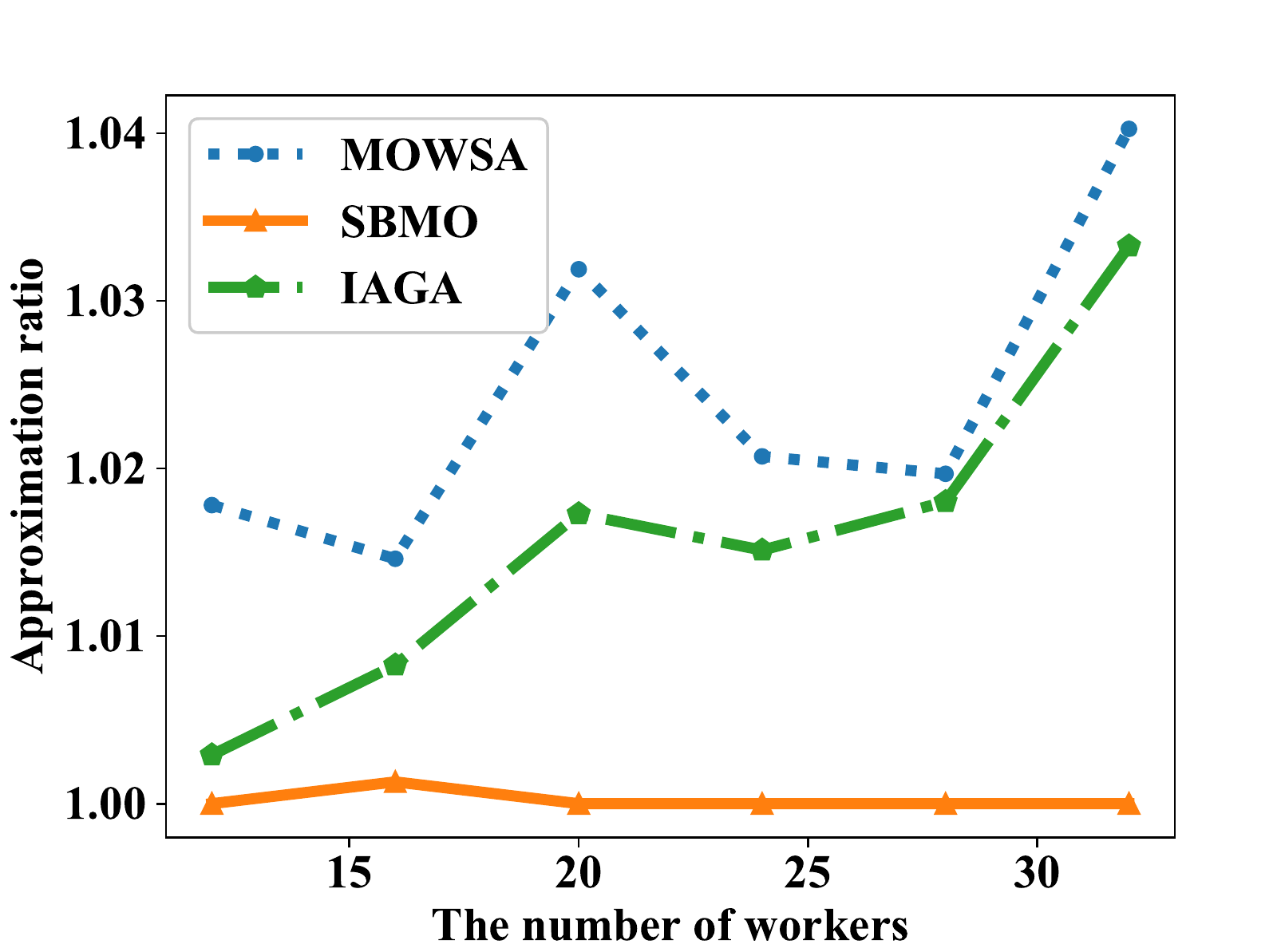}
}
\subfigure[6-stage cases]{
\includegraphics[width=2.22 in]{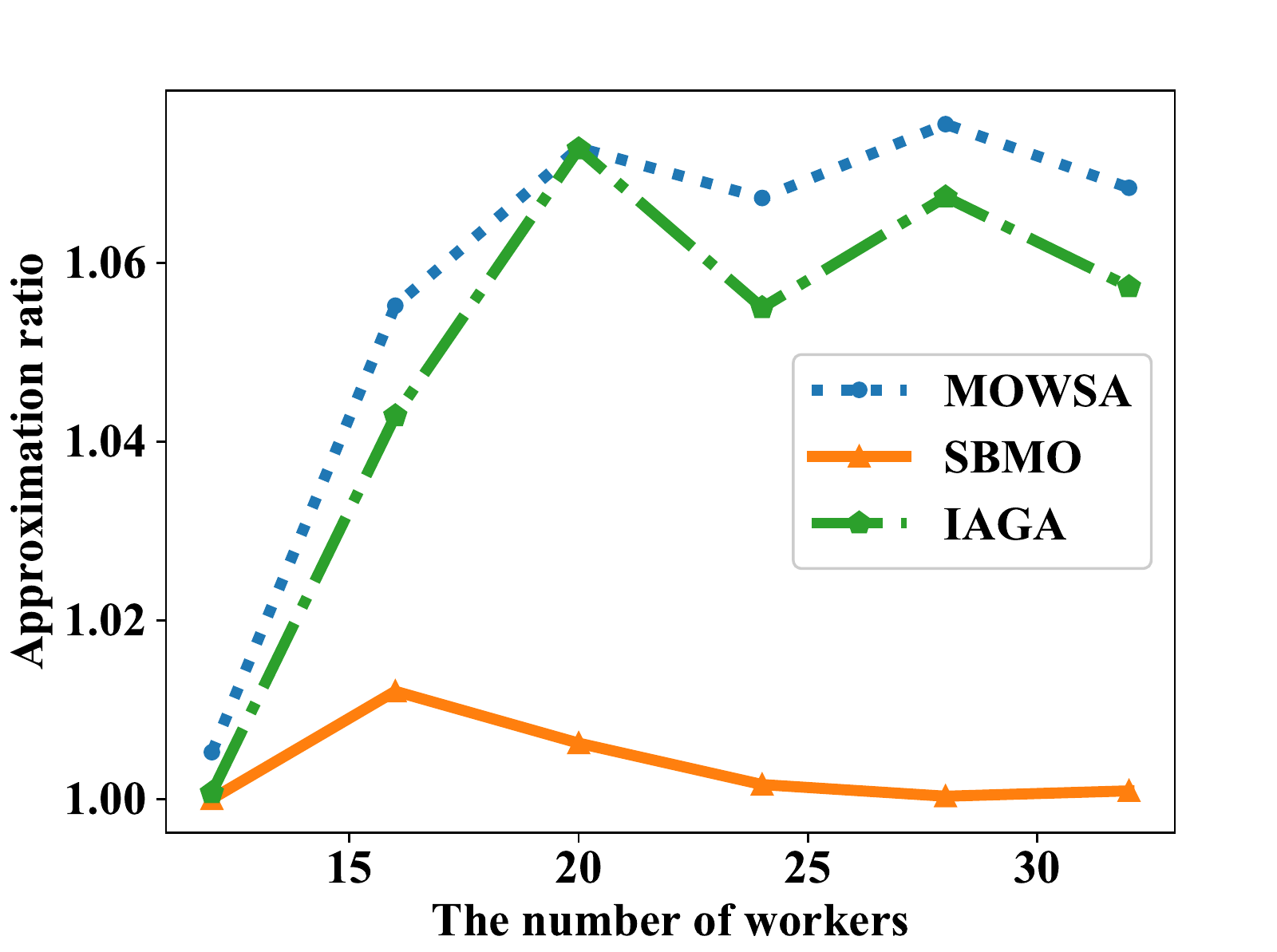}
}
\subfigure[8-stage cases]{
\includegraphics[width=2.22 in]{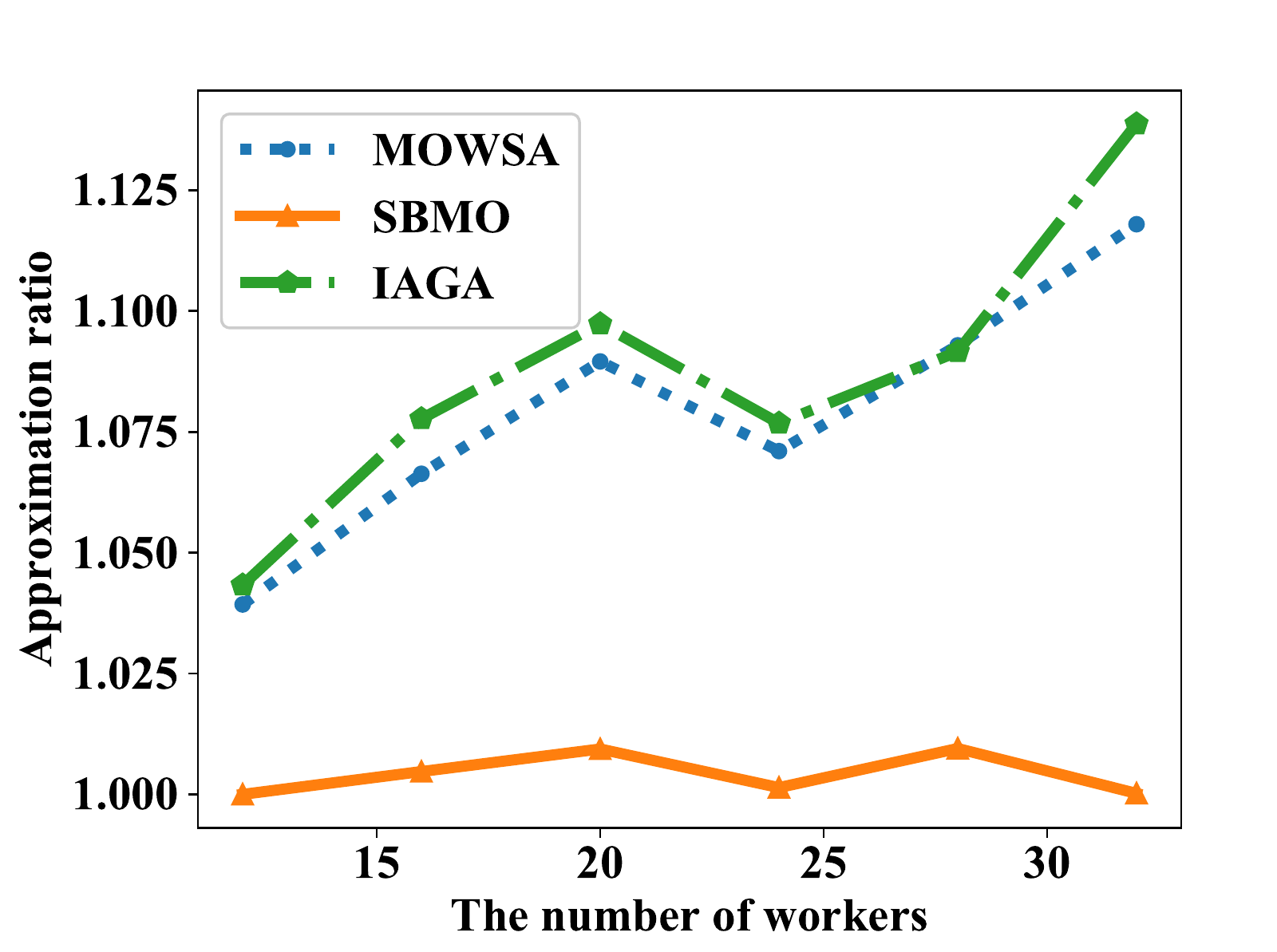}
}
\subfigure[10-stage cases]{
\includegraphics[width=2.22 in]{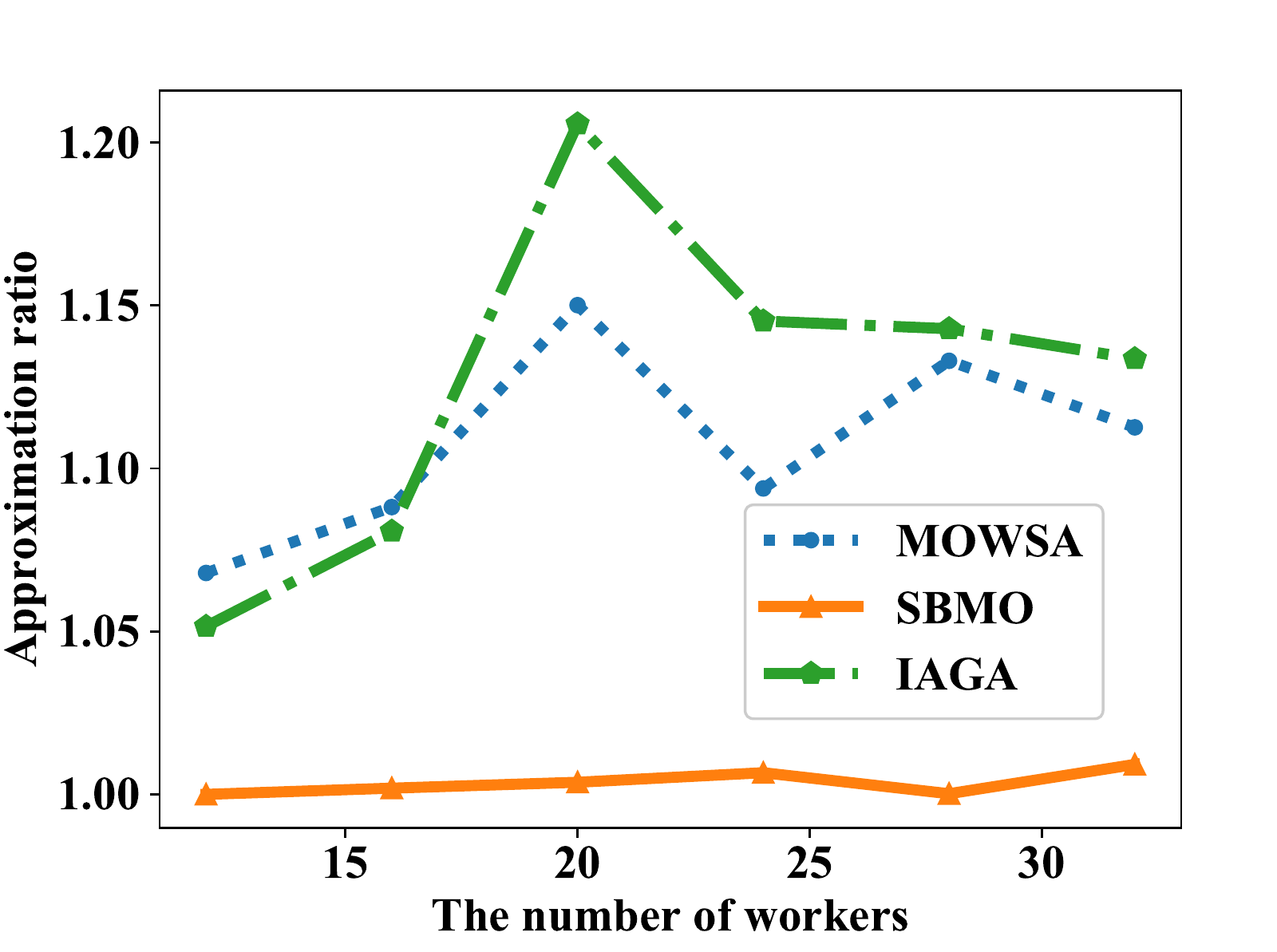}
}
\subfigure[12-stage cases]{
\includegraphics[width=2.22 in]{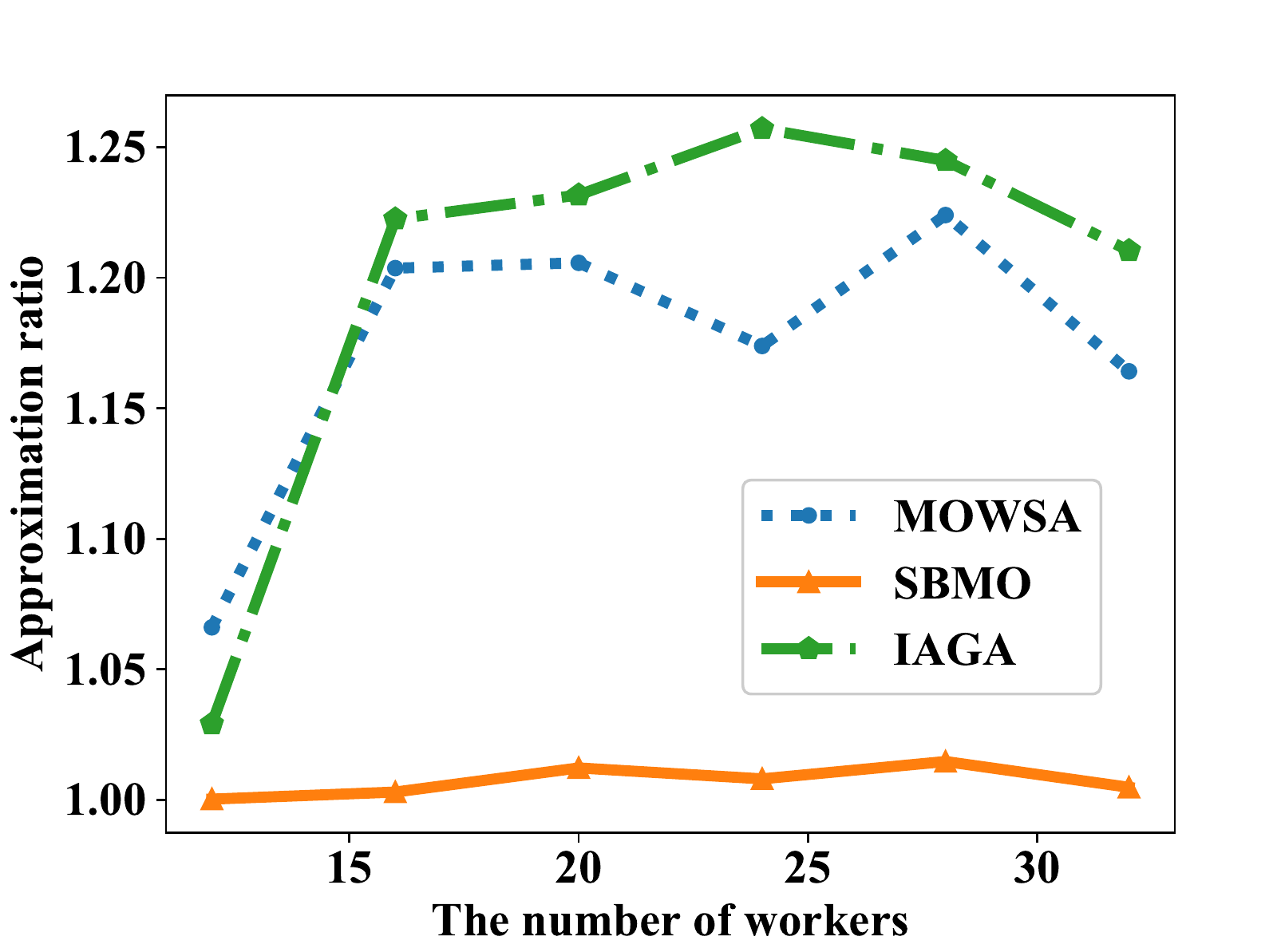}
}
\centering
\caption{The impact of $R$ on $\gamma$}
\label{AR curves}
\end{figure*}

\begin{figure*}[htbp]
\centering

\subfigure[4-stage cases]{
\includegraphics[width=2.22 in]{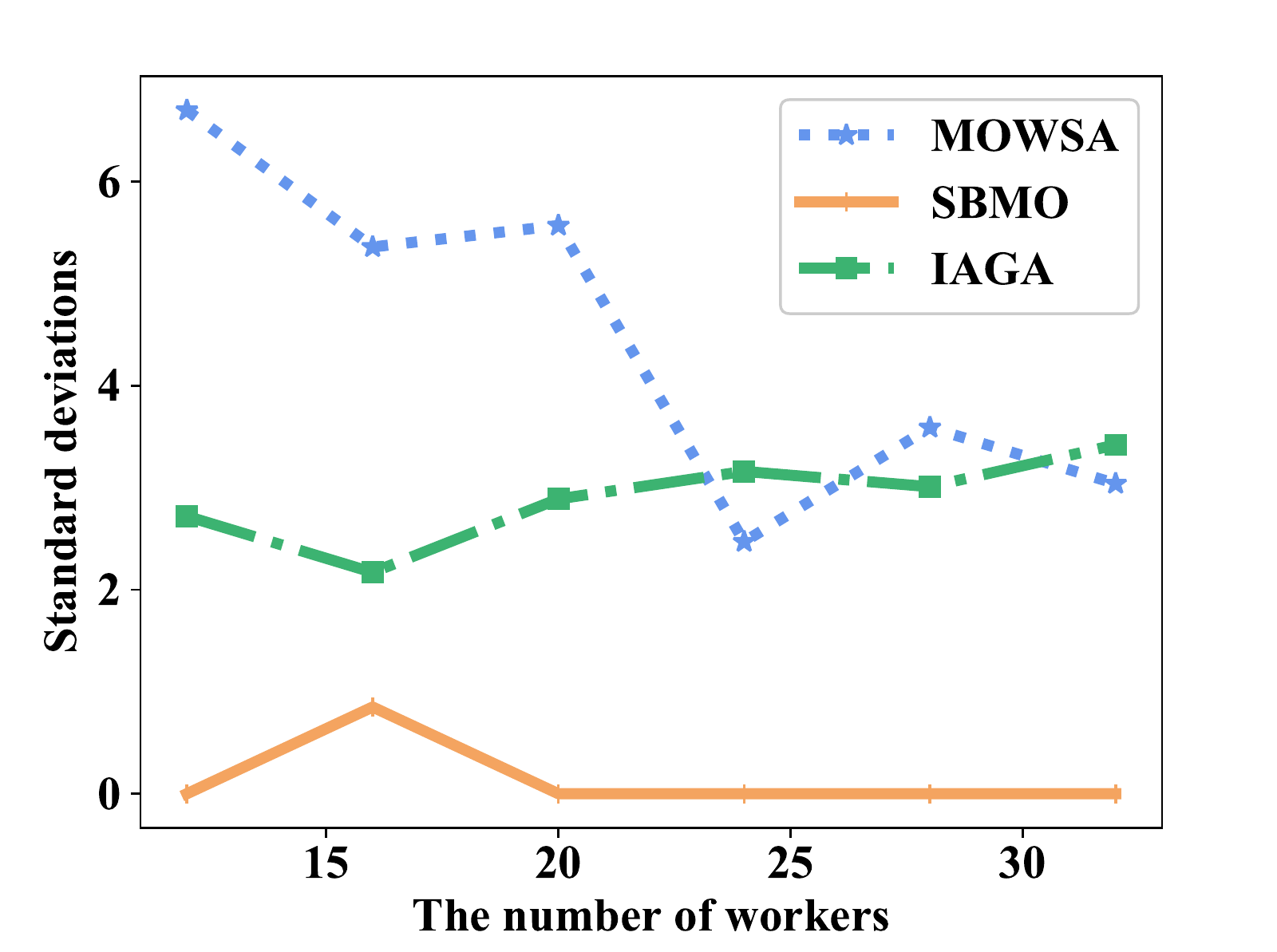}
}
\subfigure[6-stage cases]{
\includegraphics[width=2.22 in]{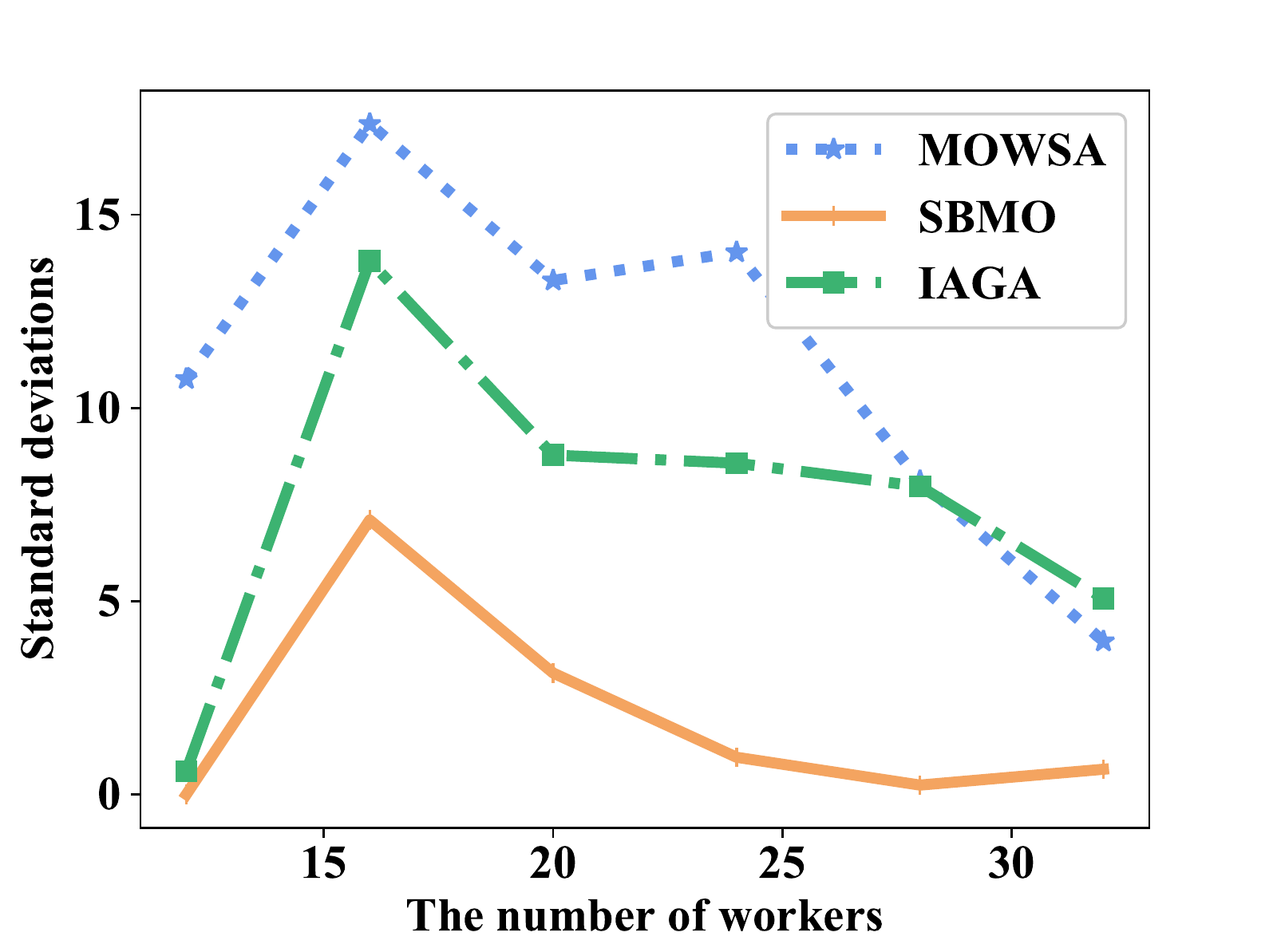}
}
\subfigure[8-stage cases]{
\includegraphics[width=2.22 in]{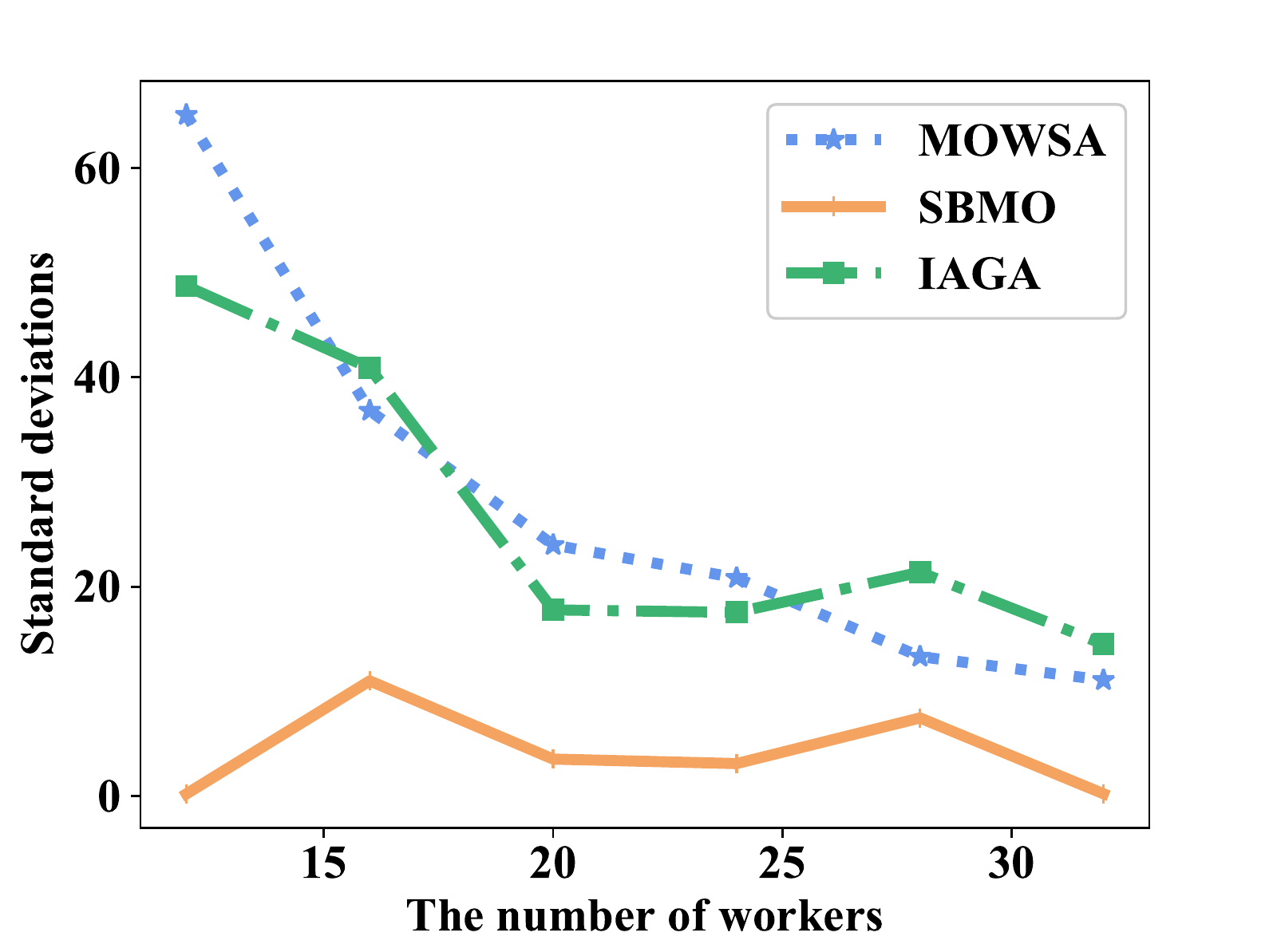}
}
\subfigure[10-stage cases]{
\includegraphics[width=2.22 in]{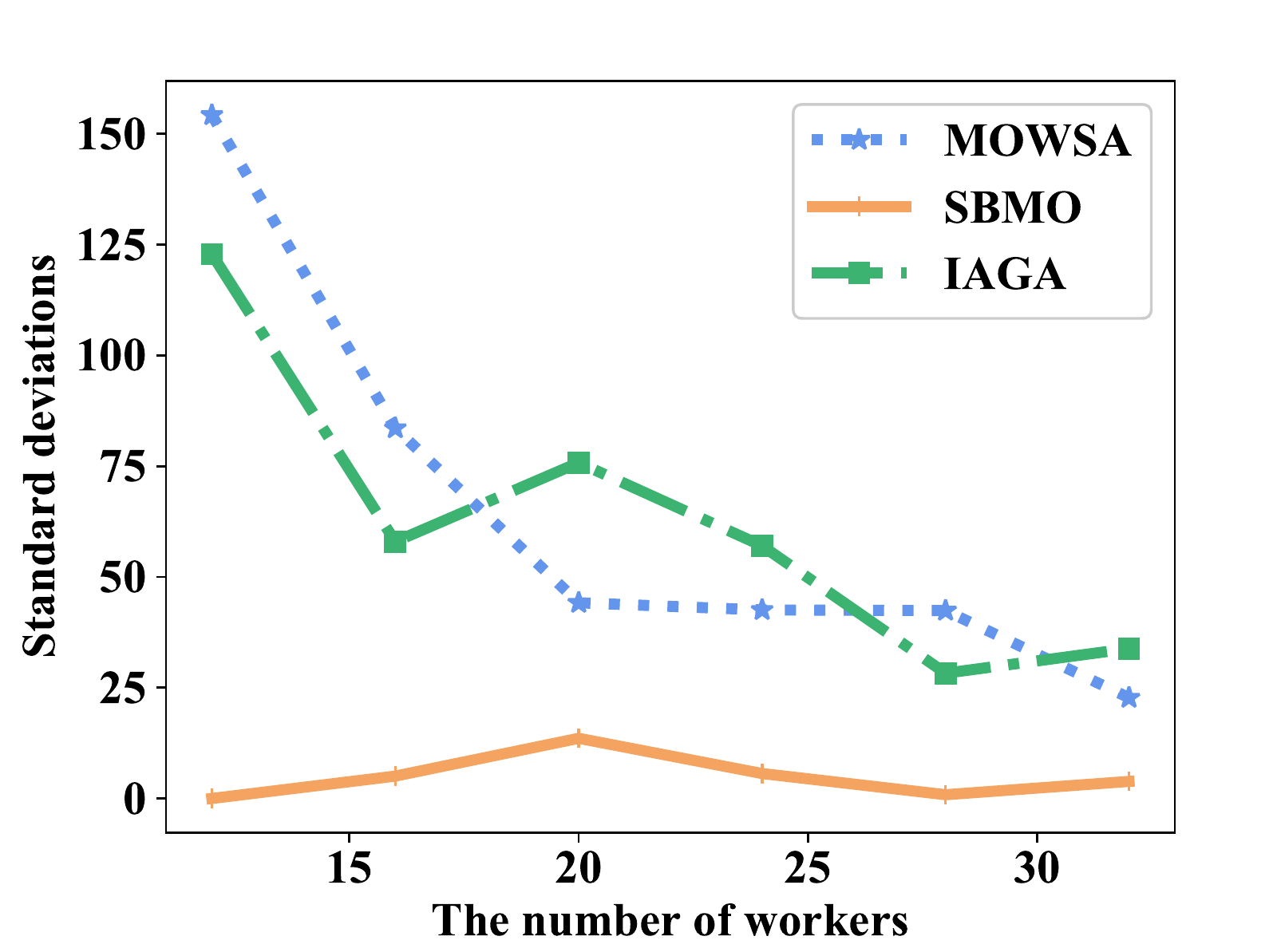}
}
\subfigure[12-stage cases]{
\includegraphics[width=2.22 in]{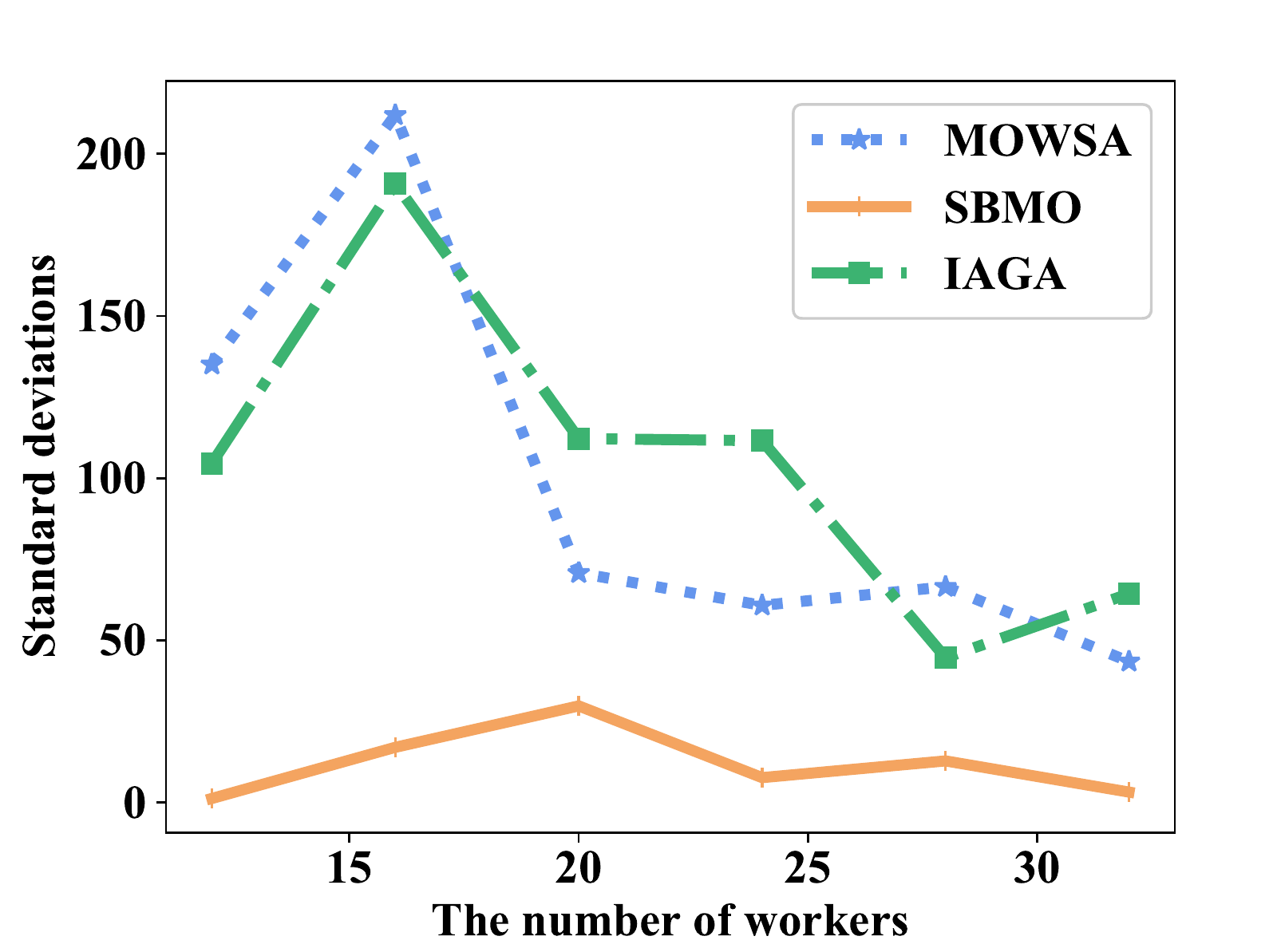}
}
\centering
\caption{The impact of $R$ on $SD$}
\label{SD curves}
\end{figure*}

\begin{figure*}[htbp]
\centering

\subfigure[4-stage cases]{
\includegraphics[width=2.22 in]{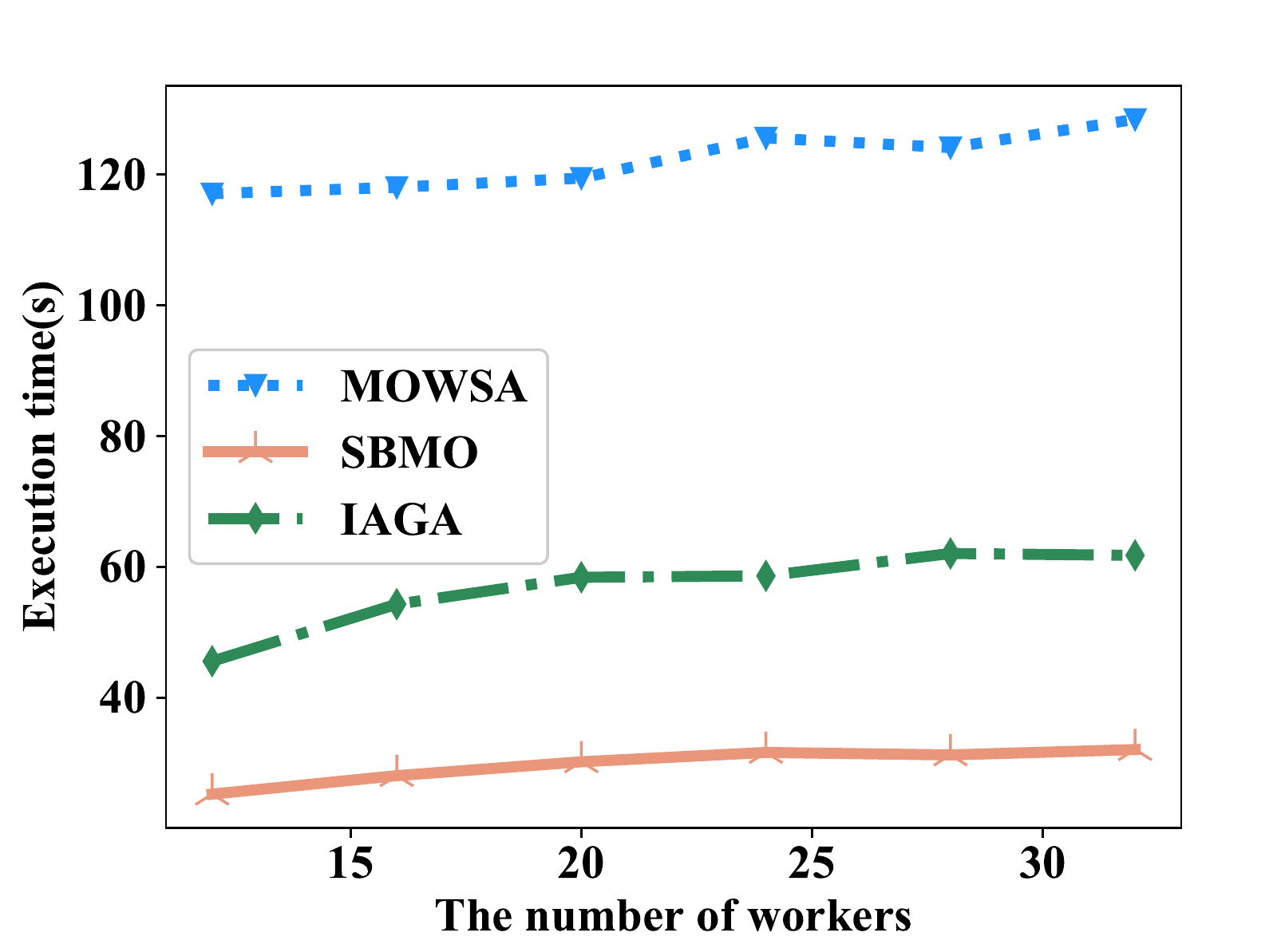}
}
\subfigure[6-stage cases]{
\includegraphics[width=2.22 in]{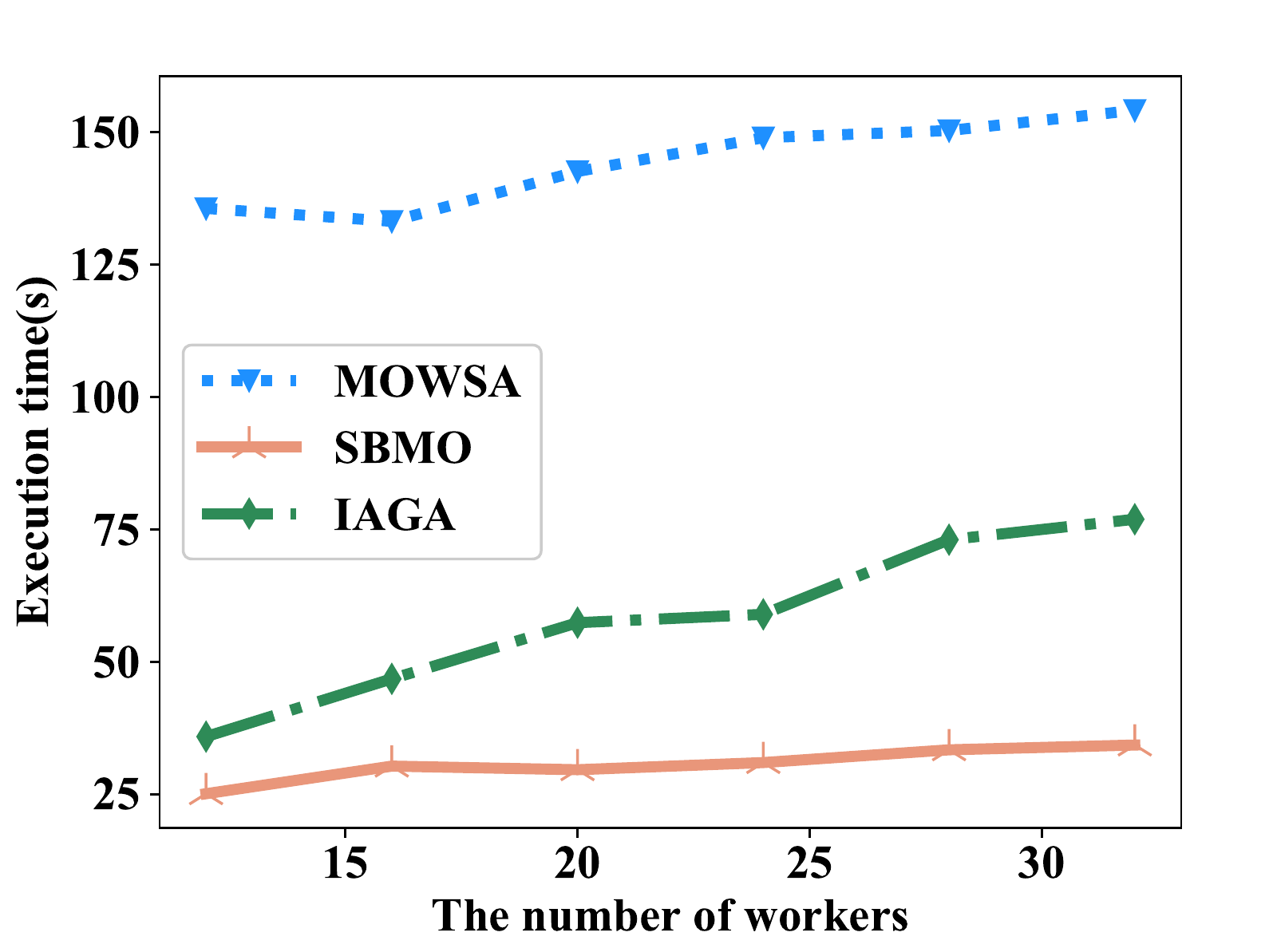}
}
\subfigure[8-stage cases]{
\includegraphics[width=2.22 in]{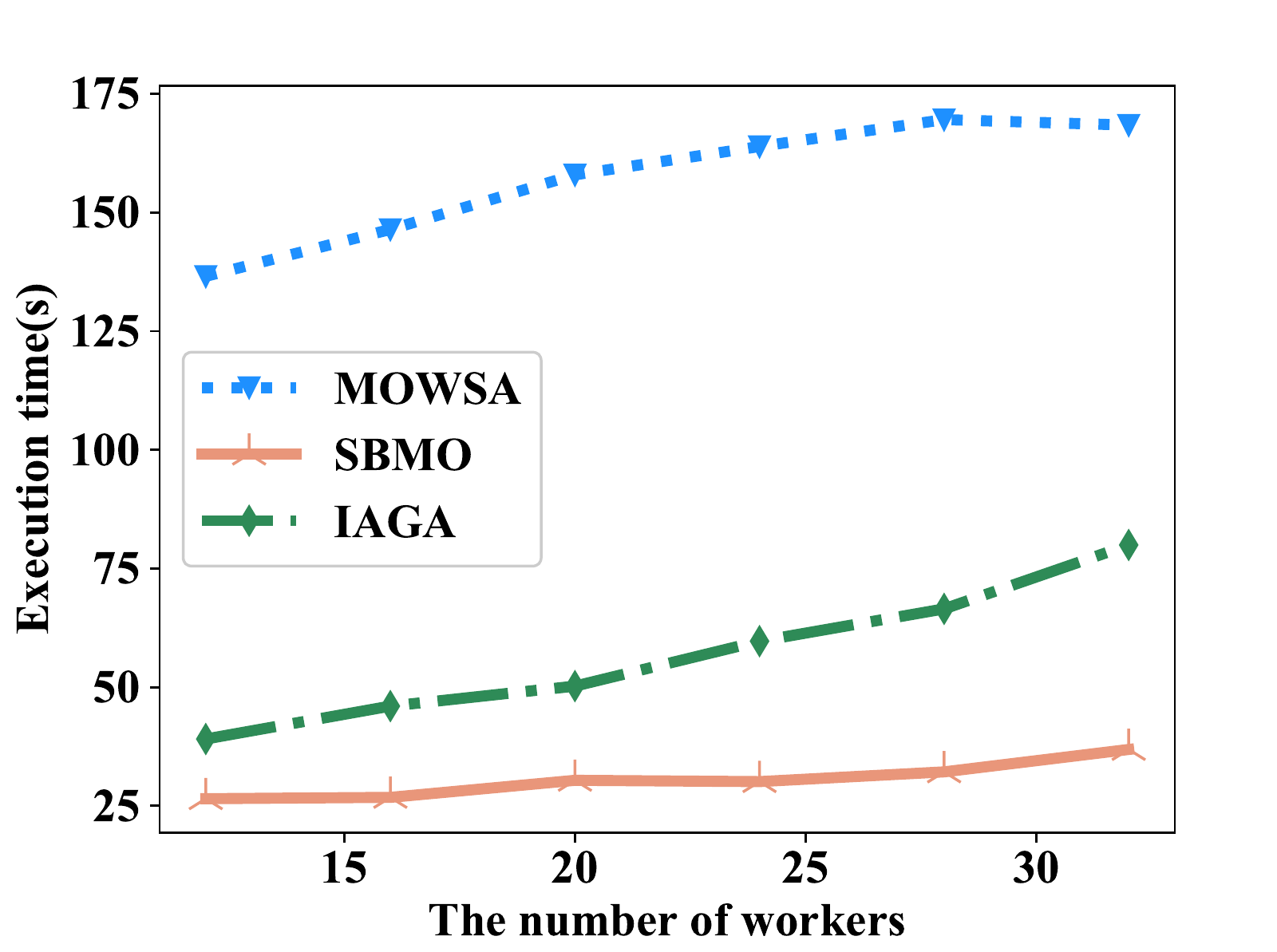}
}
\subfigure[10-stage cases]{
\includegraphics[width=2.22 in]{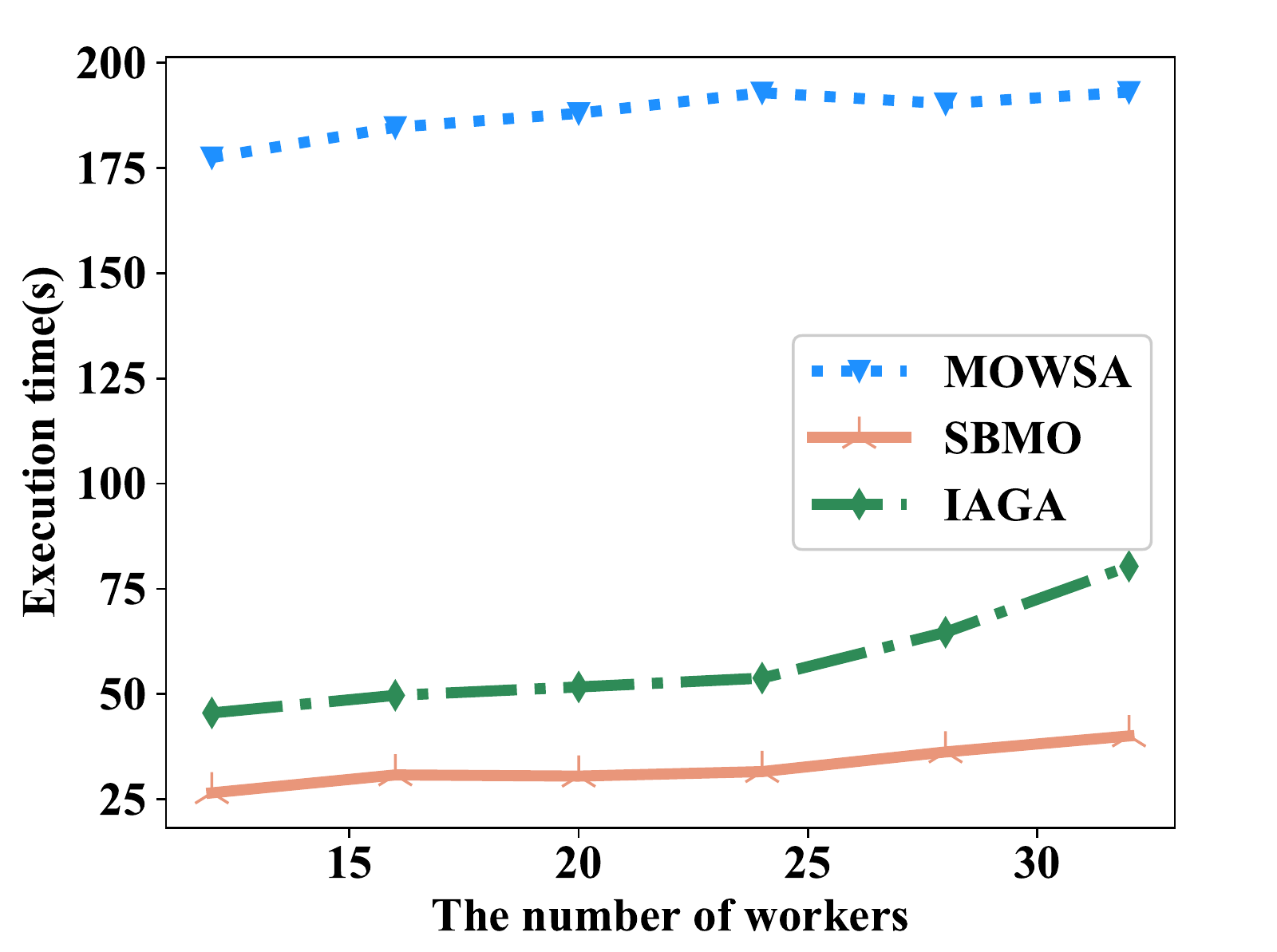}
}
\subfigure[12-stage cases]{
\includegraphics[width=2.22 in]{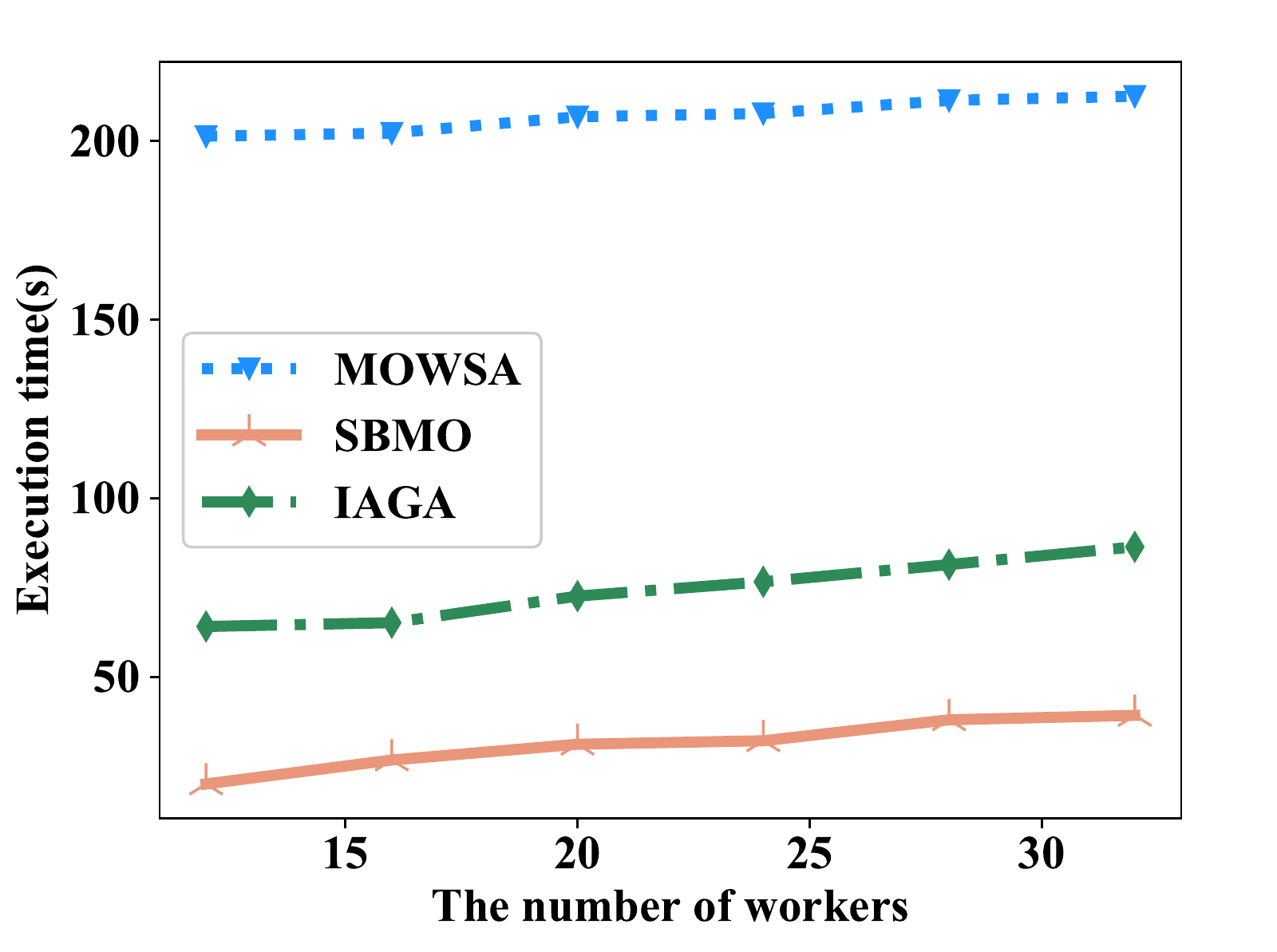}\label{time curves}
}
\centering
\caption{The impact of $R$ on execution time}
\label{Time curves}
\end{figure*}

\begin{figure*}[htbp]
\centering

\subfigure[The case of 14 workers and 6 stages]{
\includegraphics[width=3.4 in]{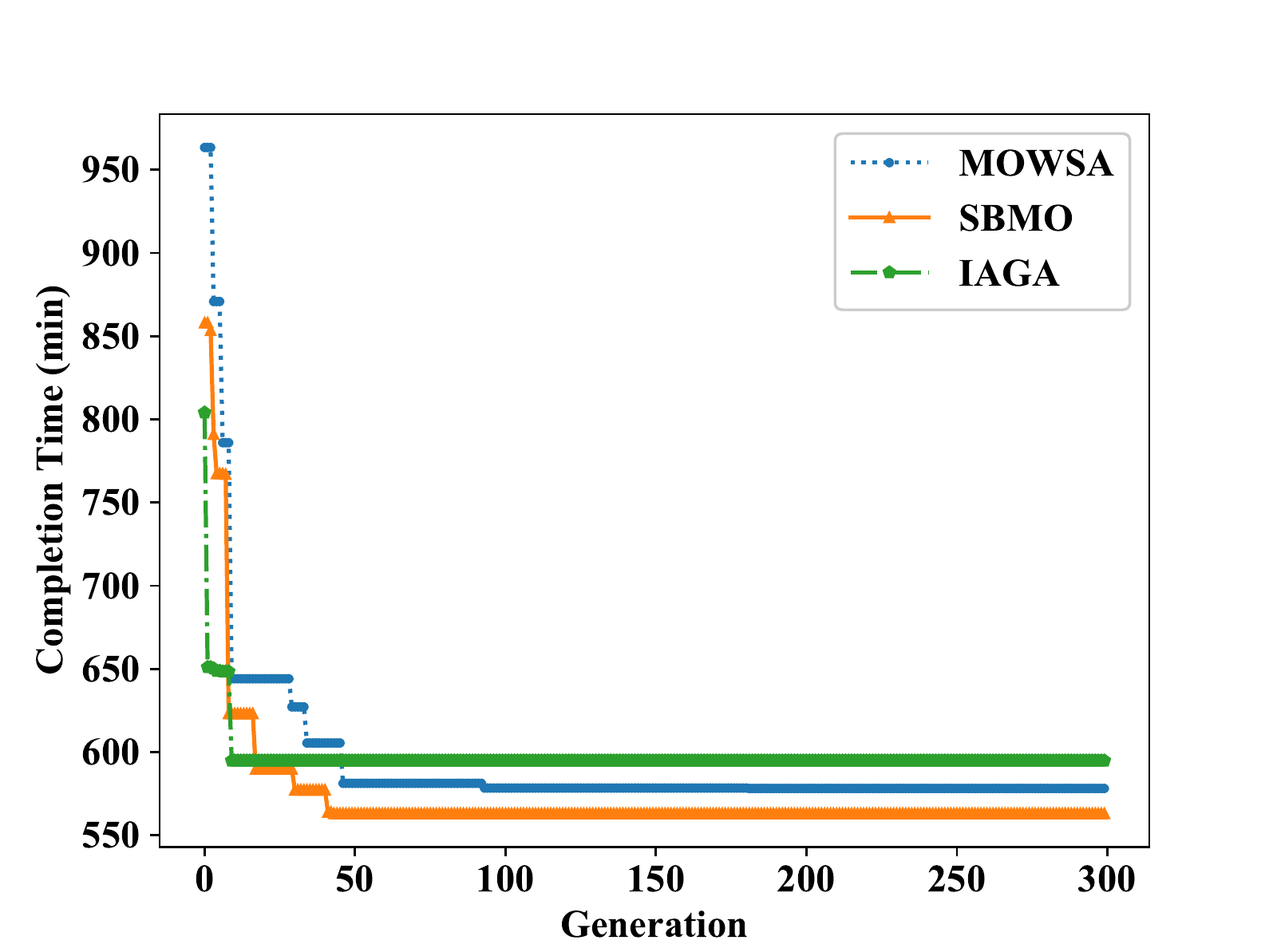}
}
\subfigure[The case of 35 workers and 10 stages]{
\includegraphics[width=3.4 in]{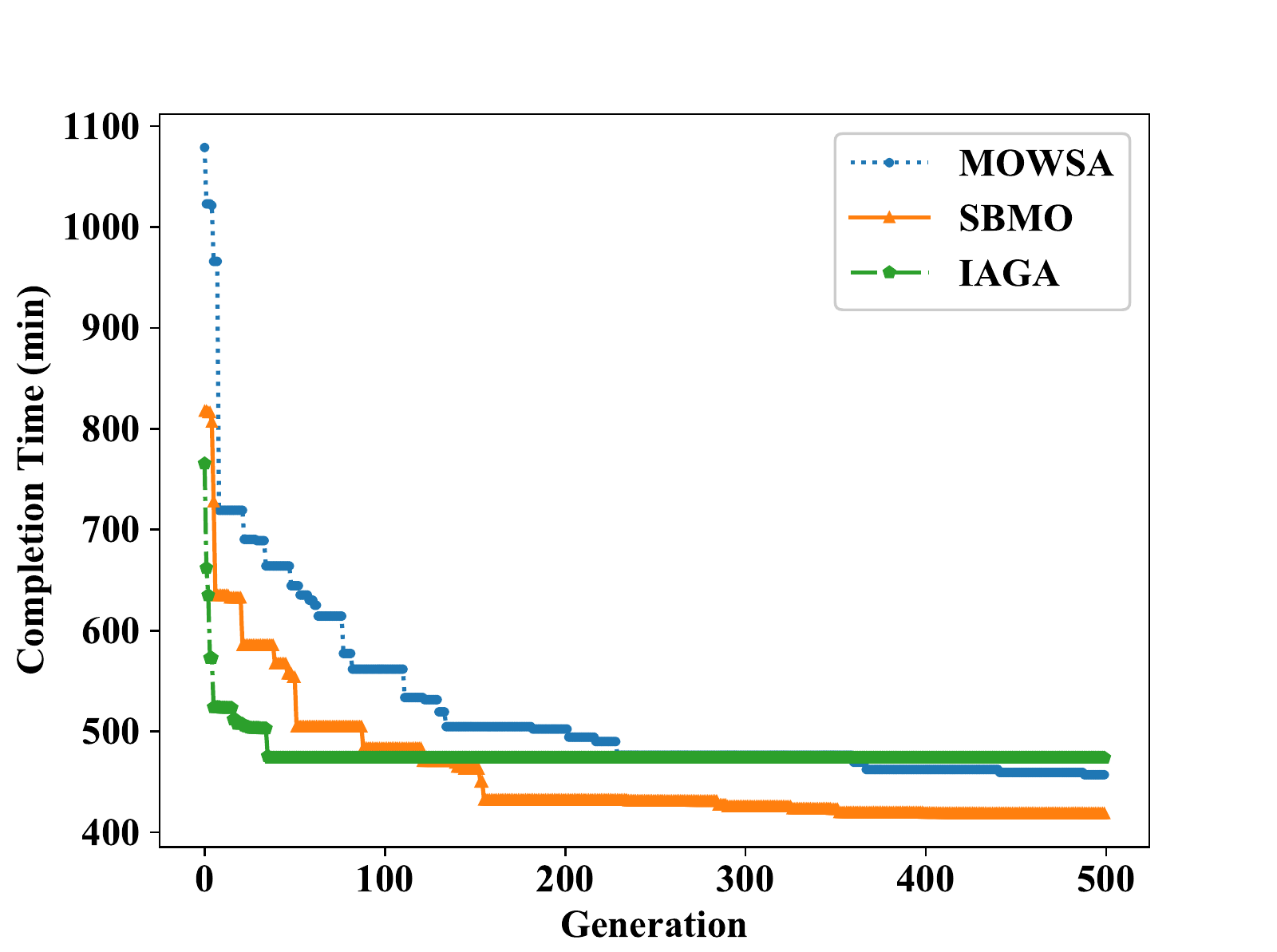}
}

\subfigure[The case of 20 workers and 12 stages]{
\includegraphics[width=3.4 in]{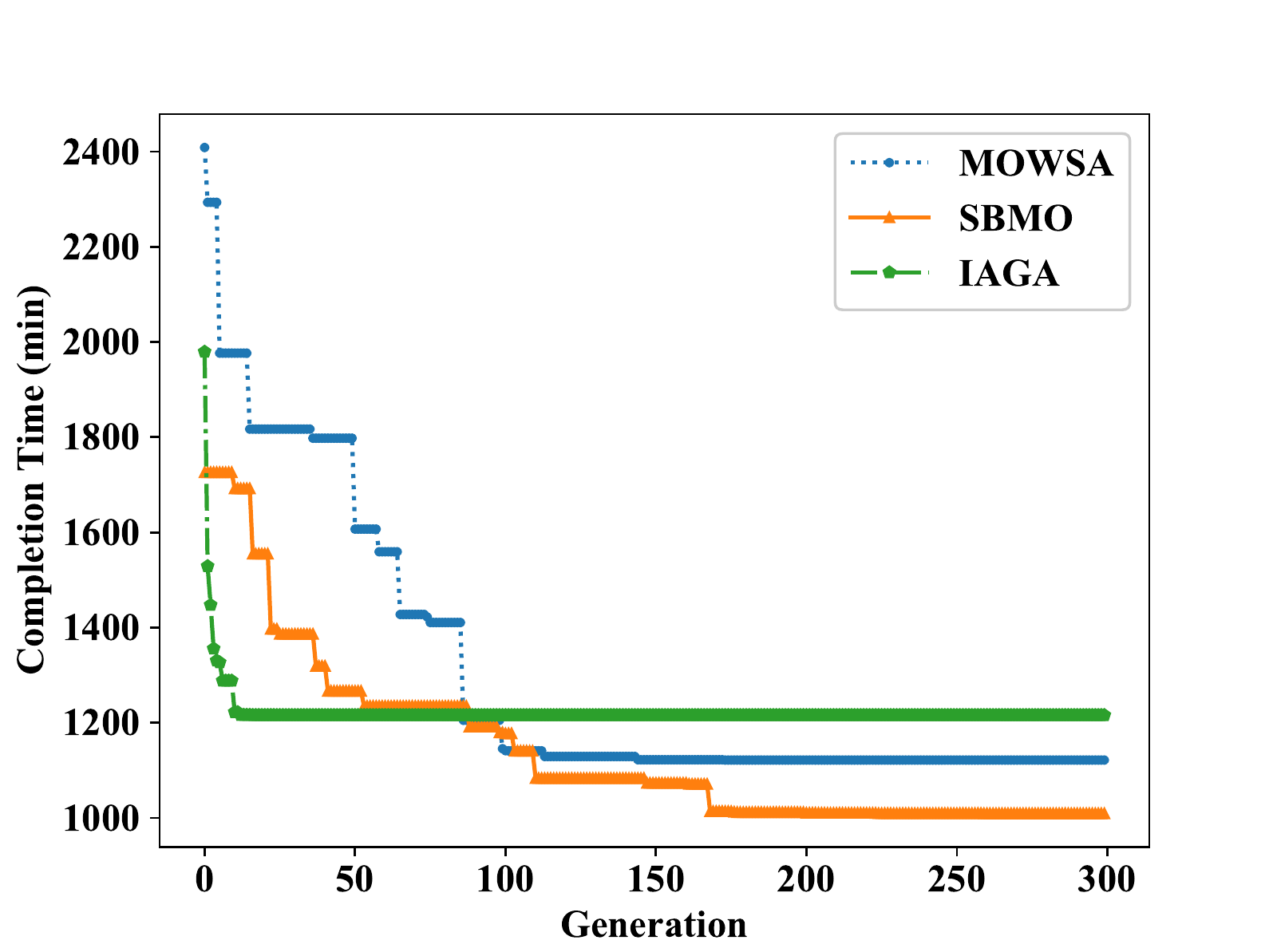}
}
\subfigure[The case of 30 workers and 12 stages]{
\includegraphics[width=3.4 in]{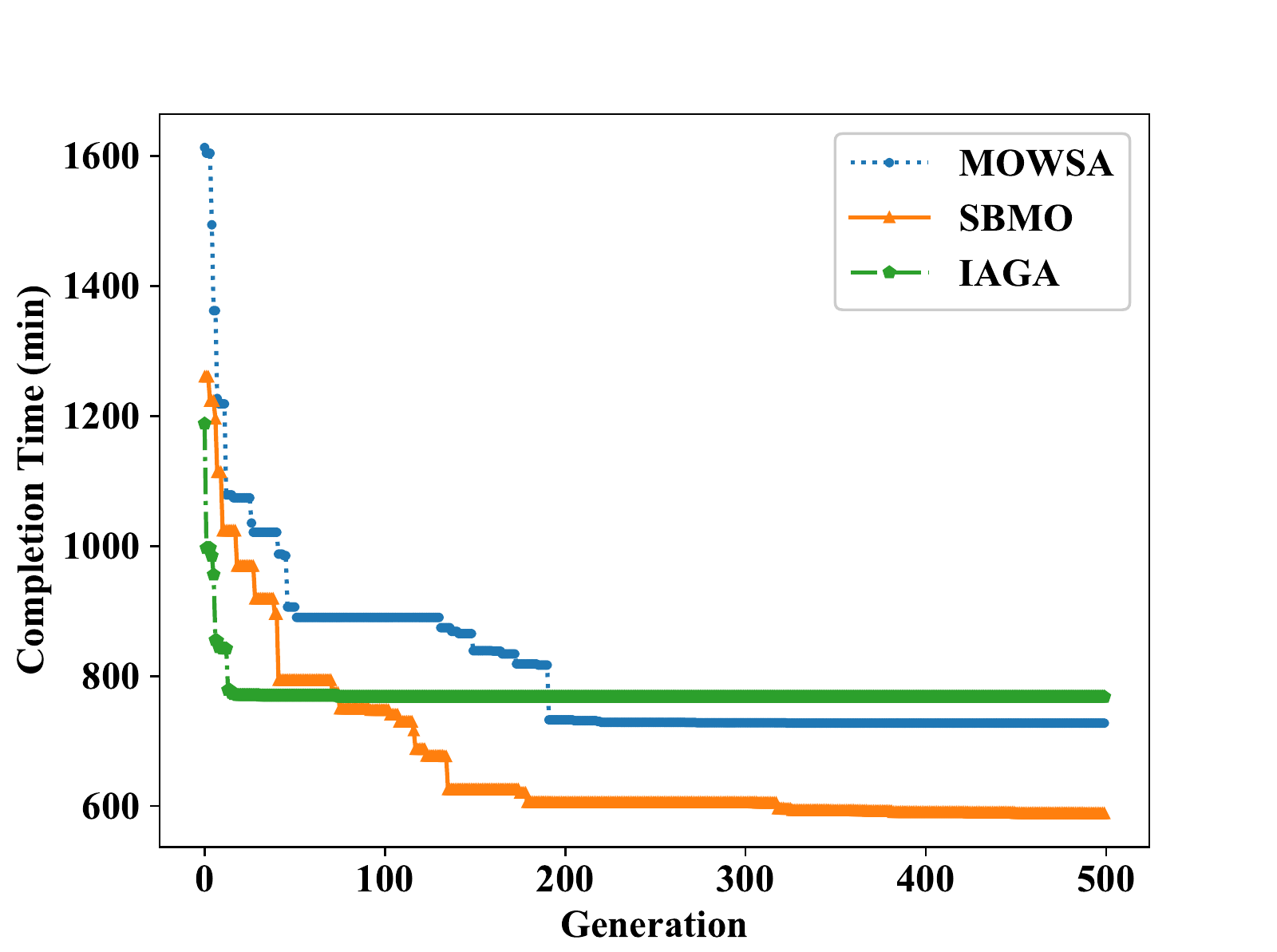}
}

\centering
\caption{Convergence curves}
\label{generation2}
\end{figure*}

\subsection{Case study}
The completion time is mainly influenced by the characteristics of workers, such as worker size and their proficiency, as well as the properties of stages, like stage scale and the unit time.
The workers' proficiency ($k_{ij}$) is uniformly distributed over (0, 1), and the unit time is a random positive number.
$R$ is varied from 12 to 32 with the increment of 4, and $N$ from 4 to 12 with the step of 2.
$D$ has little impact on the $T$, and we set $D=100$ in our simulations.

\subsubsection{Effectiveness of the neighborhood search}
We first evaluate the performance of the neighborhood search.
We compare our SBMO algorithm with the one without the module of neighborhood search, noted as SBMO-WN.
The comparison results are given in Table \ref{table1}, including 30 cases.
The first column indicates the case scale with the amounts of stages and workers.
The $2^{nd}-3^{rd}$ columns represent the comparison of $\gamma$, while the following $4^{th}-5^{th}$ columns indicates the one of $SD$.
The last two columns represent the execution time of two compared algorithms.
All 30 cases show that the SBMO has achieved better results.
The solutions obtained by SBMO are more stable by their $SD$ comparison.
Although SBMO has a little cost on the execution time, it is acceptable in practice for better solutions.

Next, we investigate the convergence speed of the comparison, shown in Fig. \ref{generation1}.
We pick up two cases with a small scale (20 workers) and a large one (30 workers) to detect the impact of the case scale.
The results verify that SBMO-WN converges prematurely regardless of the case scale.
For the case of 20 workers, SBMO-WN converges at about 150 iterations with an inferior solution, while SBMO converges at about 170 generations with a superior one.
For another case of 30 workers, SBMO-WN also converges earlier than SBMO with a poor solution.

\subsubsection{Comparison of SBMO with other algorithms}
Furthermore, we compare SBMO with the classic IAGA and popular MOWSA, which were presented for similar problems.
The analysis is given in detail from three perspectives: $\gamma$, $SD$, and execution time.

\paragraph{Evaluation $\gamma$}
As can be seen from Fig. \ref{AR curves}, the performance of SBMO dominates the best one for all cases.
In addition, Fig. \ref{AR curves} also demonstrates $\gamma$ keeps stable for the SBMO algorithm with the increasing case scale, while $\gamma$ rises in fluctuation for the other compared algorithms.
Hence, it is concluded that the case scale has little impact on the performance of SBMO, rather than the ones of IAGA and MOWSA.

\paragraph{Evaluation $SD$}
From Fig. \ref{SD curves}, it is observed that SBMO has more robust stability since IAGA and MOWSA have a higher probability of falling into local optimum.
It can be further seen that $SD$ decreases with the rising number of workers with the same stage case.
It is because that the adjustment of schedule becomes less impact on $T$ with the expanding workers' size.

\paragraph{Evaluation of computation efficiency}
MOWSA adopts an extra similarity calculation for diversity maintenance.
Hence, the running time of MOWSA is much longer than that of SBMO and IAGA.
For specifically, the time complexity of MOWSA and IAGA is $O(GQ^2R)$ and $O(GQR)$, respectively.
Although the time complexity of IAGA is the same as the one of SBMO, IAGA has the extra operations of tournament selection and elite retention.
Thus, it is more time-consuming than SBMO.
As shown in Fig. \ref{Time curves}, SBMO has the highest computation efficiency.
Therefore, the simulation results match the theoretical analysis.
The execution time enlarges monotonically with the increase of workers' scale.

\paragraph{Evaluation of convergence}
To verify the effectiveness of SBMO, we also give the convergence curves of these three algorithms.
Fig. \ref{generation2} shows that the convergence speed increases with the expanding case scale.
We can find that IAGA converges very quickly at the beginning, but it is easy to fall into a local optimum.
MOWSA performs better than IAGA because MOWSA has an advantage in maintaining diversity.
However, it also has the problem of premature convergence.
In summary, SBMO can obtain higher-quality solutions, regardless of the case scale.

\section{Conclusion}
Focused on the impact of manpower, we first define a new problem, FSMSP, for the shop schedule.
To handle the FSMSP considering the proficiency of workers, we establish a pure integer nonlinear programming model.
Furthermore, we propose a Self-encoding BMO algorithm to minimize the completion time.
In addition, we design a neighborhood search scheme based on the characteristics of FSMSP to solve the problem of local optimum.
Eventually, both theoretical analysis and extensive simulations demonstrate that the proposed SBMO algorithm achieves high computational efficiency, superior approximation ratio, and robust stability, compared with the current others.

\section{Data Availability Statement}
The data that support the findings of this study are available from the corresponding author, WQX, upon reasonable request.

\end{document}